\pdfoutput=1
\documentclass{article}

    \usepackage[preprint]{neurips_2022}

\usepackage[utf8]{inputenc} % allow utf-8 input
\usepackage[T1]{fontenc}    % use 8-bit T1 fonts
\usepackage{hyperref}       % hyperlinks
\usepackage{url}            % simple URL typesetting
\usepackage{booktabs}       % professional-quality tables
\usepackage{amsfonts}       % blackboard math symbols
\usepackage{nicefrac}       % compact symbols for 1/2, etc.
\usepackage{microtype}      % microtypography
\usepackage{xcolor}         % colors
\usepackage{ulem}
\usepackage{multirow}
\usepackage{amsmath}
\usepackage{caption}
\usepackage{subfigure}
\usepackage{graphicx}
\usepackage{threeparttable}

\newcommand{\specialcell}[2][c]{%
\begin{tabular}[#1]{@{}c@{}}#2\end{tabular}
}

\newcommand{\qmethod}{BiFeat}

\newcommand{\yanc}[1]{\textcolor{blue}{\textbf{Feng: #1}}}
\newcommand{\yuxin}[1]{\textcolor{orange}{\textbf{Yuxin: #1}}}

\title{{\qmethod}: Supercharge GNN Training via Graph Feature Quantization}
\author{%
  Yuxin Ma\\
  University of Science and Technology of China\\
  \texttt{leeadama@mail.ustc.edu.cn}\\
  \And
  Ping Gong\\
  University of Science and Technology of China\\
  \texttt{gpzlx1@mail.ustc.edu.cn}\\  
  \And  
  Jun Yi \\
  University of Nevada\\
  \texttt{junyi@nevada.unr.edu} \\
  \And  
  Zhewei Yao \\
  Microsoft\\
  \texttt{zheweiyao@microsoft.com} \\  
  \And  
  Cheng Li\\
  University of Science and Technology of China\\
  \texttt{chengli7@ustc.edu.cn}\\  
  \And  
  Yuxiong He \\
  Microsoft\\
  \texttt{yuxhe@microsoft.com} \\   
  \And  
  Feng Yan \\
  University of Houston\\
  \texttt{fyan5@central.uh.edu} \\   
}

\begin{document}

\maketitle

\begin{abstract}
    Graph Neural Networks (GNNs) is a promising approach for applications with non-Euclidean data.
    %, such as recommendation system, knowledge graph and computational biology. 
    However, training GNNs on large scale graphs with hundreds of millions nodes is both resource and time consuming.
    Different from DNNs, GNNs usually have larger memory footprints, and thus the GPU memory capacity and PCIe bandwidth are the main resource bottlenecks in GNN training.
    % Host-to-Device copy of graph feature is the bottleneck of GNN training throughput and host memory size restricts the graph scale. But
    To address this problem, we present {\qmethod}: a graph feature quantization methodology to accelerate GNN training by significantly reducing the memory footprint and PCIe bandwidth requirement so that GNNs can take full advantage of GPU computing capabilities. 
    Our key insight is that unlike DNN, GNN is less prone to the information loss of input features caused by quantization. 
    %with minor accuracy degradation. 
    We identify the main accuracy impact factors in graph feature quantization and theoretically prove that {\qmethod} training converges to a network where the loss is within $\epsilon$ of the optimal loss of uncompressed network. %features.   %to measure how high compression ratio a graph can tolerate.
    We perform extensive evaluation of {\qmethod} using several popular GNN models and datasets, including GraphSAGE on MAG240M, the largest public graph dataset.
    The results demonstrate that {\qmethod} achieves a compression ratio of more than 30 and improves GNN training speed by 200\%-320\% with marginal accuracy loss. In particular, {\qmethod} achieves a record by training GraphSAGE on MAG240M within one hour using only four GPUs.
    
\end{abstract}

\section{Introduction}

Graph Neural Networks (GNNs) is a promising approach for modeling the structural information of data. Popular GNN models such as GraphSAGE\cite{hamilton2017inductive}, GCN\cite{kipf2016semi}, and GAT\cite{velivckovic2017graph} have recently achieved state-of-the-art (SOTA) performance in a broad range of fields, such as social network\cite{wu2020graph}, knowledge graph\cite{xu2019cross}, recommender system\cite{hu2020gpt}, and bioinformatics\cite{chereda2021explaining}. 
%\yanc{please add references.} \jy{Done.} 

However, training GNNs on large graphs  (e.g.,
graphs with hundreds of millions of nodes) is non-trivial because of the expensive computation and memory demands during the training process. Deep network structure~\cite{liu2020towards,chen2020simple} benefits GNN performance, but is also more demanding in computing and memory resources, especially when deployed in GPUs where the memory capacity is usually limited. 

GNN training has a unique data pipeline.  First, GNNs inherits the irregular data processing flow of graph analytics as the input nodes feed to the network are randomly sampled. Second, the computing pattern is regular and fast as the forward, backward, parameter update operations are the same across iterations and the computing is fast due to the relatively small number of parameters of GNNs compared to Deep Neural Network (DNNs). 
%\yanc{can we be more specific what is irregular and regular here? %\jy{irregular graph processing running on regular neural network}}
The dynamic and irregular data accesses combined with the regular and fast computation lead to frequent data loading, which causes poor CPU/GPU utilization as the computation is often waiting for data feeding. %Therefore, the training is slow and expensive. 

To tackle these challenges, we propose {\qmethod}, a graph feature quantization method that accelerates GNN training by reducing the memory footprint and PCIe bandwidth requirement of graph features without compromising the training accuracy. 
Our key observation here is that GNNs features have much higher tolerance to quantization error compared to DNNs, thanks to the averaging during neighbor information aggregation, where errors can cancel out.
%can cancel out most of the error by averaging. 
%(we also find not only mean aggregator, learnable aggregators(like GCN and LSTM) would also learn to product this cancel out).
%} \yanc{please briefly summarize the reason here -- the average option?}
To better understand the performance impact of graph feature quantization, we theoretically prove the convergence of {\qmethod} with a statistical bound on loss compared to the uncompressed network training and give the key factor.

In summary, we make the following main contributions:
\begin{itemize}
\item We make an initial effort on accelerating GNN training by performing feature quantization, and identify the main performance impact factors in graph feature quantization.
\item We propose a graph feature quantization method {\qmethod} and make theoretical analysis on its convergence. {\qmethod} in general and can be used with any GNN models.
%We further theoretical prove that  feature compression in GNNs can keep the original accuracy.
%\item {\qmethod} is decoupled with models, so general GNNs models can benefit from it.
\item Extensive experimental evaluation using various GNN models and datasets demonstrate that {\qmethod} achieves a compression ratio of more than 30, which improves GNN training speed by 50-200\% with less than 1\% accuracy loss. In particular, {\qmethod} achieves a record by training GraphSAGE on MAG240M within one hour using only four GPUs.
%show that with Quantization aware training, {\qmethod} can provide 2x-3x speedup in large scale graphs, even on small-scale graph, it also demonstrates the advantages.
\end{itemize}
\section{Background and Motivation}

\subsection{GNN Mini-batch Training}
%\subsection{GNN mini-batch training}
Graph Neural Networks (GNNs) extracts graph structure information by aggregating other nodes' features.
%However, there do exists some problems making use of GNNs in real production scene. 
Production-scale graphs are usually very large with a giant number of nodes and edges, as well as the associated features. For instance, as shown in Table~\ref{tab:data_statistics}, the MAG240M graph dataset consists of more than 240 million nodes and 3.4 billion edges, where each node has 768 features. Its total size is more than 349 201.7 GB, largely exceeding the GPU memory space (up to 80 GB) of the most advanced, commercially available GPU~\cite{stone2010gpu}. Therefore, it is impossible to fit the graph structure and features into the GPU memory for doing full graph training. 

To address the limited GPU memory challenge, GraphSAGE\cite{hamilton2017inductive}, GCN-based FastGCN\cite{chen2018fastgcn}, and ClusterGCN\cite{chiang2019cluster} do not use full graph training, instead, introduce mini-batch training for GNNs.  
%make it able to train GNNs on GPU.
The core idea of GNN mini-batch training is to use a sampler to extract a subgraph from the full graph at each step and load the subgraph on GPU for training. We have been witnessing that it has been widely adopted, and become the native training method supported by the mainstream GNN training frameworks such as DGL\cite{wang2019deep} and PyG\cite{fey2019fast}. Though being able to train GNNs over large graphs, the mini-batch training suffers a serious data loading problem. This is because, during the training process, the structure information of the sampled subgraph and the corresponding features need to be continuously moved from CPU memory to GPU memory via PCIe, which is very slow due to the mismatch between the limited PCIe bandwidth and the large volume of transferred data. This would lead to waste of the expensive GPU resources. We demystify the data loading problem below. %and is impossible to do full graph training on it.
%%CL Data loading problem, how serious it is.
%GNNs over using the subgraph.

\begin{figure}[ht]
\begin{minipage}[t]{0.465\linewidth}
\centering
\includegraphics[width=\textwidth]{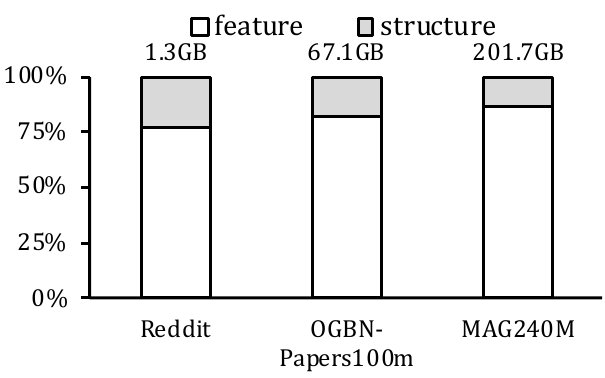}
     \caption{The ratios of feature and structure sizes of different datasets}
     \label{fig:feature_percentage}
\end{minipage}
\hfill
\begin{minipage}[t]{0.465\linewidth}
\centering
\includegraphics[width=\textwidth]{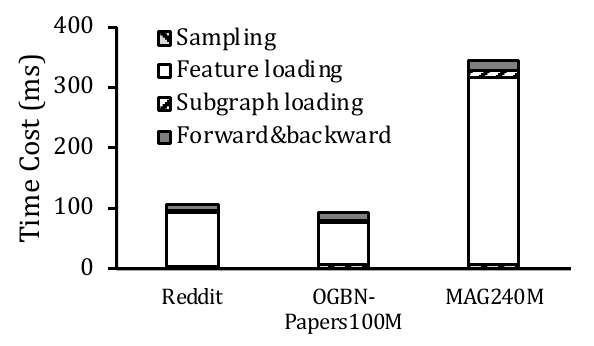}
     \caption{Breakdown of each phase of the mini-batch training of GraphSAGE on three datasets}
     \label{fig:original_breakdown}
\end{minipage}
\end{figure}

\subsection{Motivation}

To conduct an in-depth analysis of the data loading pipeline, %%CL  of GNN mini-batch training, 
we train GraphSAGE~\cite{hamilton2017inductive} on Reddit, OGBN-Papers100M, and MAG240M datasets, with the training framework DGL. The statistics of these datasets can be found in Table~\ref{tab:data_statistics}. We summarize our key findings of below. 
%We summarize the unique data access patterns of GNNs below. \yanc{we need to say more clearly why the patterns below are related to the motivation of the feature compression approach.}
\begin{itemize}
    \item Compared to the structure information, as shown in Figure~\ref{fig:feature_percentage}, the feature data dominates the size of a graph dataset. For instance, for the largest graph MAG240M, its feature occupies 174 GB, accounting for 86.3\% of its total size. %%CL (see Figure~\ref{fig:feature_percentage}) and is accessed multiple times (hundreds of times for GraphSAGE) during an epoch. 
    %Thus, reducing the feature data size can greatly help relieve GPU memory and PCIe bandwidth pressure.
   
    \item The feature data loading is the bottleneck. Specifically, as illustrated in Figure~\ref{fig:original_breakdown}, across the three training tasks, the feature loading time is 81\%, 74\%, and 91\% of their training time costs, respectively. Furthermore, GNN models usually have a relatively small number of parameters compared to DNNs, thus the computation time of forward and backward propagation is very small compared to feature data loading. This also makes the feature data loading a very frequent task and difficult to be hidden by computation. 
    \item In each iteration, input nodes are randomly sampled from a large set of nodes. Such randomness makes it challenging to utilize the spatial-temporal locality in data access. Thus the help from caching is usually limited.
    %random access makes it inefficient to store data in hard drive like CNN and ordinary DNN tasks.
    %\item Even though different nodes have different probability to be sampled, the variance of probability is relatively small due to the density of graph. Thus small cache usually has a poor cache hit rate. \yanc{the logic here is not very clear.}\yuxin{data access frequency do have different between nodes, but the difference isn't very big, hot data is not that hot}
\end{itemize}

The above patterns indicate that the GNN feature data loading is a random yet heavy workload.
Due to the large scale of GNN feature data (hundreds of GB to TB), it is infeasible to store all raw GNN feature data in GPU memory.
Storing features in CPU memory can address the limited GPU capacity issue but also causes a data movement bottleneck due to the frequent and heavy data loading. 
Thus, reducing the feature data size can greatly help relieve GPU memory and PCIe bandwidth pressure.
%When training GNN using GPUs, such data movement bottleneck can still be observed even when using the CPU memory as the bandwidth of PCIe does not match the fast GPU training speed. 
%Figure[figure] shows the time breakdown of different tasks during the training and it is clear that the data loading takes most of training time.
%Therefore, storing features directly in GPU memory is necessary.  
%and even CPU memory is not the best option, ideal case we hope features are stored in GPU memory.
%However, due to the large scale of GNN feature data (hundreds of GB to TB), it is infeasible to store raw GNN feature data in GPU memory. 
%and it would require a extremely large memory size to store it. the host memory to gpu data transfer is limited by PCIE bandwidth, causing low efficiency of training. 

%\yanc{we need a paragraph to talk about why feature compression is a promising approach.}\jy{Added.}
To reduce the feature data size, compression is a natural solution. 
The common compression techniques include sparsification~\cite{lin2017dgc, aji2017sparse} and quantization~\cite{strom2015TBQ, wen2017terngrad, seide20141onebit}. %%CL \pingc{cite gradient compression here} 
For spasification, most graphs, especially large ones, have features in the form of float numbers, which is dense. %%CL Tiny graphs like cora and citeseer have sparse and even one-hot features. But in most of these situation, data loading is not the problem, and 
Unlike this, quantization is more general, and can handle dense features. %%CL \yanc{we need a short justification why we use quantization instead of sparsity.}
Therefore, our goal is to devise a new GNN feature quantization methodology to compress feature data so that the data volumes transferred between CPU memory and GPU memory have been greatly reduced. For quantization, our key insight is that 
%there is significant redundancy in feature data. 
GNNs features have much higher tolerance in reduced precision compared to DNNs as errors can be cancelled out by averaging during neighbor information aggregation. In this paper, we will provide both theoretical and practical results to support this observation. %%\yanc{do we have any results to support this?}
%Since various types of features usually result in information redundancy as well as increase of neural network nodes which would further lead to model training difficulty and over-fitting, efficient and light-weight feature compression is no-doubt expected to perform in large and complex system like GNNs. Feature compression technique embraces encoding process, and followed by decoding process, to relieve the burden of communication between nodes while keeping the accuracy level. However, the applications of deep learning feature compression are hindered by that deep learning models are normally designed and trained for specific tasks, and the top-layer features are extraordinary abstract and task-specific, making such compressed features difficult to generalize. 

\subsection{Related Work}

Here, we briefly iterate a few most relevant works that share the same goal of addressing the slow data pipeline challenge in GNNs as we do. PaGraph\cite{lin2020pagraph} utilizes spare GPU memory to cache frequently visited graph data to speed up data loading. However, due to the large GNN feature size, the sampling randomness, and the limited GPU memory, spare GPU memory can only cache a small portion of features, especially for large datasets. Therefore, such method does not scale well for large graphs, and observes diminished cache hit ratios and speedups with the increase in the graph size.
% But with quantization, we can cache all the feature of largest seen datasets(MAG240M), avoiding feature host2device copy.

GNNAutoScale\cite{fey2021gnnautoscale} supports large scale GNN training by using historical embedding instead of re-calculating neighbor's embedding at every step. This reduces GPU memory consumption and data loading time since less number of nodes are involved at each step. However, historical embedding increases CPU memory cost by a factor of the number of layers, results in poor scalability for large-scale datasets.  
% may be benefit from our BiFeat?
Global sampling\cite{dong2021global} aims at solving the data loading problem by using a small cache in GPU with higher sampling priority for nodes in cache.  
However, such method may lead to accuracy degradation and is also coupled with specific models. The modified forwarding method is also slower than the original one and thus limits the actual speedup. 
%, its modified forwarding method costs more time than original one, makes the speedup not very ideal.
Existing works, such as Binary Graph Neural Networks\cite{bahri2021binary},  VQ-GNN\cite{ding2021vq}, Sgquant\cite{feng2020sgquant}, Degree-quant\cite{tailor2020degree}, also use quantization for GNNs. However, they focus on reducing the size of weights and activations, which is orthogonal to our work as we aim at addressing the data loading bottleneck by compressing the GNN features. 
%so computation is speeded up but data loading remains a bottleneck, and they do not help reduce host memory requirement, they also cause accuracy degradation.

\if 0
 \item In GNN mini-batch training, there are significant overlap in nodes between subgraphs, so the overall data amount loaded into GPU is significantly larger than the whole graph size. \yanc{what is the conclusion here? use full graph training is actually preferred because less data movement?}\yuxin{full graph training can't be run on giant graph due to hardware limitations} 
 \fi

%%CL \input{motivation}
\section{GNN Feature Quantization}
\label{sec:quantization}
%By quantizing the bitwidth for the whole network, data quantization method, one of the main and efficient compression technology, is widely used to overcome the computation burden of modern Neural Network models. Quantization can have potentially very high compression ratio especially for large-scale graphs processing system like GNNs. In our work, we define compression ratio as the division of uncompressed size of original data and compressed data size: $Compression \  Ratio(CR) = \frac{Uncompressed\ Size}{Compressed\ Size}$, where our target is to achieve a high ratio result to better take advantage of system bandwidth. 

Quantization has been widely used to reduce model size~\cite{ding2021vq, hu2016BNN, liu2020BRN} and gradients~\cite{wen2017terngrad}, 
%researched and specifically designed to compress models in order to speed up computation and reduce memory and storage cost, 
but applying it to GNN features has not been explored before. 
Given compressing feature data is different than compressing model or gradients, it is non-trivial to find an appropriate quantization method and understand the corresponding accuracy impact, compression benefits, and computation cost. %%CL is a highly non-trivial request question.  

%\subsection{{\qmethod}}
In this paper, we propose {\qmethod}: a GNN feature quantization methodology that supports both Scalar and Vector quantization approaches to accelerate GNN training by significantly reducing the memory footprint of GNN feature data. We first explore the Scalar quantization method (in Section~\ref{subsec:SQ}), which is simple and compute-efficient. But its flip-side is the limited compression ratio of up to 32. To enable GNN training with extremely large graphs, in Section~\ref{subsec:VQ}, we additionally investigate the validity of the vector quantization method~\cite{gray1984vector}, which can raise the compression ratio up to a few hundreds or thousands. However, this comes at the price of compute-intensive feature preprocessing. We give the implementation details of the two methods in Section~\ref{sec:quantization}. Following that, we give a theoretical analysis to prove their negligible impacts on training convergence and accuracy in Section~\ref{sec:analysis}. Finally, we run intensive experiments with state-of-the-art GNN models and various graph datasets to validate the effectiveness and efficiency of applying {\qmethod} to boost GNN mini-batch training over large graphs.
%\yanc{can we be more specific how we used the methods}
%{\qmethod} leverages quantization to reduce memory footprint and data loading cost. It

%Next, we introduce and compare two commonly used quantization methods: Scalar quantization and Vector quantitation. %, and our proposed method {\qmethod}.
%it has been not used to compress the input features so far as we know. The reason is, in most DNNs, input quantization causes huge accuracy degradation and low speedup (due to data loading is not the main bottleneck in most cases)

%Despite this non-negligible drawback in compression method, we find GNNs can tolerate much higher compression ratio than normal DNNs. It is due to its aggregated neighbor information, the quantization error of different nodes would cancel out during aggregation. Meanwhile, compressing feature by quantization can save  the limited memory space (especially for GPU) and speed up training tremendously. Next, we will illustrate the two most common quantization methods: Scalar quantization and Vector quantitation, and our proposed method {\qmethod}.

\subsection{{\qmethod}-SQ}
\label{subsec:SQ}
{\qmethod} Scalar quantization ({\qmethod}-SQ) projects continuous value to several discrete values, thus we can use a low bit width integer to indicate a float number~\cite{haase2020dependent}. {\qmethod}-SQ may cause a rounding error proportional to the gap between two discrete values.
The commonly used uniform quantization method uniformly selects these values so that the gaps are even and the maximum rounding error is reduced. %%CL\yanc{please change the description to GNN feature specific.}
However, using it on graph features requires a different setup. Most large-scale graphs have features near normal distribution because features are usually outputs generated from models such as BERT\cite{devlin2018bert}. Since most of the values are near zero, to reduce the overall rounding error,
%and to narrow down the rounding error, 
we use logarithm based quantization method:
%\yuxin{not sure if we need to propose this, since we almost only use 1 bit scalar quantization}\jy{change it to have instead of propose}

\begin{equation}
Q(x)  =  \begin{cases}
2^{k-1}-1 - \lfloor  \frac{Clip(log_{2}(-x))-e_{min}}{e_{max}-e_{min}} *2^{k-1}\rfloor , &  x<0   \\
2^{k-1} + \lfloor \frac{Clip(log_{2}x)-e_{min}}{e_{max}-e_{min}} *2^{k-1}\rfloor, &  x\ge0     \\
\end{cases}
\label{eq:quantization}
\end{equation}
where \textit{x} is the original feature, $e_{min}$ and $e_{max}$ are the minimum and maximum values respectively of binary logarithm on non-negative $x$ after clipping outlier values, i.e., $Clip(log_{2}|x|)$. This is actually a uniform quantization of the clipped logaritm. We tradeoff some dynamic range to reduce the overall rounding error.   %influenced by $x$. 
Dequantization is similar in a reversed manner:
%operation of quantization as we just mentioned.
\begin{equation}
Q^{-1}(q)  =  \begin{cases}
exp2 \left (\frac{(2^{k-1}-0.5 - q)*(e_{max}-e_{min})}{2^{k-1}}+e_{min} \right ), q<2^{k-1}   \\
exp2\left (\frac{(q-2^{k-1}+0.5)*(e_{max}-e_{min})}{2^{k-1}}+e_{min} \right ), q\ge 2^{k-1}     \\
\end{cases}
\label{eq:dequantization}
\end{equation}
where $q$ is the quantized value from equation \ref{eq:quantization}. 

Scalar quantization can compress and decompress quickly, but it has a maximum compression ratio of 32 when $k=1$ and the sign of x is used for the quantized value (positive or non-positive). Such quantization method only slightly degrade the accuracy. We don't differentiate zero and negative values for the compatibility of discrete original value like in one-hot code\cite{rodriguez2018beyond} and binary features\cite{heinly2012comparative}. For example, on tiny graph datasets like Cora\cite{bojchevski2017deep} and Citeseer\cite{kipf2016semi}, quantizing these datasets would not bring much information loss. 

\subsection{{\qmethod}-VQ}
\label{subsec:VQ}
{\qmethod} Vector quantization ({\qmethod}-VQ) views all GNN feature data as vectors and uses clustering methods such as kmeans to group features into clusters. It uses the clustering centers to present the values in the cluster~\cite{gersho2012vector}. These centers are stored in a codebook and the index can be used to indicate each value. {\qmethod}-VQ method can achieve a much higher compression ratio because it represents multiple float values with an integer.
{\qmethod}-VQ is usually slower than {\qmethod}-SQ in quantization, though the dequantization speed is similar. 

However, due to the GPU memory limitation, it is very difficult to directly quantize vectors with hundreds of dimensions through a large codebook with thousands of entries. Features need to be split into multiple parts with less dimensions, each part has a independent codebook.
So the number of codebook entries and the length of each entry are two parameters in {\qmethod}-VQ, namely length and width. 
Besides these two parameters, another setting is using Euclidean distance or cosine similarity to measure the distance between vectors. These settings are critical for achieving a high compression ratio with little accuracy impact.
Our experiments show that cosine similarity consistently performs better. This is because the relative size of different dimensions is more important. And less partition combined with large codebooks helps too.%%CL \yuxin{may need polish here} 

{\qmethod}-VQ requires careful hyperparameters tuning, but can possibly reach a higher compression ratio. 
If the original feature bitwidth be $b$, we can calculate the theoretical compression ratio by $ CR = width * b / log_2length$.
 Due to bit alignment, The value of length may need to be carefully selected. Codebooks are usually far smaller than the quantized features, and we see increasing the length won't cause the compression ratio to drop too much. 
%%CL \yuxin{do we need this analysis to compression ratio}
%Some performance analysis(among hyper parameters)[figure]

\subsection{System Integration}

{\qmethod} performs feature quantization against the target graph datasets before training. This is one-time job, and can be re-used across multiple runs. At every training iteration, {\qmethod} loads quantized features, corresponding to a sampled subgraph, from CPU memory to GPU memory, and recovers compressed features into their correct formats by conducting dequantization on GPU, before kicking off the forward and backward computation. The dequantization is required as we want to make {\qmethod} be transparently integrated with the GNN training framework DGL, and  keep the computation kernels unchanged with their input features with common formats. {\qmethod} significantly reduces the data loading size and thus accelerates the training. 
%to be loaded, and since we don't 
Since the data and its format loaded into the GNN model is unchanged, no modification is required in GNN models. Therefore, {\qmethod} can be easily implemented with only a few lines of code.

{\qmethod} also supports batching, which is especially useful when processing large scale graph like MAG240M. Specifically, {\qmethod} loads and dequantizes a small batch at a time to reduce the GPU memory consumption.
Next, we detail {\qmethod} Scalar and {\qmethod} Vector quantization approaches.
%one can process datasets way larger than machine memory

\section{Theoretical Analysis}
\label{sec:analysis}

We prove a graph neural network with sufficient width, starting from a random initialization and using input features with small quantization error, would eventually converge to a network where the loss suffered with input error is within $\epsilon$ of the optimal loss using original features. 

\newtheorem{theorem}{Theorem}

We denote the input features as $X = (x_1, x_2, ... , x_N) \in \mathbb{R}^{N \times d}$, here $N$ is the number of nodes and $d$ is the number of feature dimensions. The features after quantization is $X'= (x'_1, x'_2, ... , x'_N) \in \mathbb{R}^{N \times d}$ satisfying $\|x_i-x'_i\|_{2} \le \delta, \forall i \in [N]$.
\newtheorem{iid_error}[theorem]{Assumption}
\begin{iid_error}[Independent rounding error]
\label{iid_error}
The randomness of rounding error is independent among nodes and feature dimensions. 
\end{iid_error}
% \textit{Remark.}
% Independent rounding error among nodes would be cancelled out during aggregating, help reduce the impact of quantization to the final loss.
% The features are also independent among dimensions, but not necessarily independent among nodes.
% Feature of two nearby nodes may tend to be the same, so eventually .
\newtheorem{smooth_loss}[theorem]{Assumption}
\begin{smooth_loss}[Assumption on loss function]
\label{smooth_loss}
The loss function is Lipschitz-smooth and convex.
\end{smooth_loss}

% Let the rounding error satisfy Assumption \ref{iid_error} and loss function satisfy Assumption \ref{smooth_loss},
\begin{theorem}
\label{convergence}
Given $\epsilon$ > 0, $L$-layer GNN with a large enough width $m$. With quantization error satisfy Assumption \ref{iid_error} and loss function satisfy Assumption \ref{smooth_loss} If we run projected gradient descent for $T$ steps, then with high probability we have
$$
\min_{t=1,\cdot\cdot\cdot, T}(Loss(W_{t},X') - Loss(W_{*},X)  \le  \epsilon
$$
\end{theorem}
\textit{Remark.}
$L(W_{*},X)$ is the optimal loss of network trained with original features, and $L(W_{t},X')$ is the minimum loss during {\qmethod} training. So this indicates {\qmethod} training wouldn't cause much larger loss compared with using original features.

We focus on the difference between GNNs and traditional neural networks. 
With Assumption\ref{iid_error}, the aggregation phases help us bound the scale of perturbation caused by quantization. The impact of rounding error to the loss would be negligible as GNN gets deeper.

We are able to quantify this by a key factor $\hat{C}$, decided by graph structure and negatively correlated with the average $L$-hop neighbor number, which would help guide the choice of quantization setup.
 
The detailed proof of Theorem\ref{convergence} is in Appendix A.
\begin{table}[!t]
    \centering
    \small
    \caption{Statistics of graph datasets. K, M, and B stand for thousand, million, and billion, respectively.}
    \begin{tabular}{ccccccc}
        \toprule
        Datasets & Task & \#Graphs & \#Avg Nodes & \#Avg Edges & \#Labels & \#Features \\
        \midrule
        Reddit~\cite{hamilton2017inductive}          & Node  & 1 & 233.0K & 11.6M  & 41    & 602 \\
        Papers100M~\cite{hu2020open} & Node  & 1 & 111.1M & 1.6B   & 172   & 128 \\
        MAG240M~\cite{hu2020open}         & Node  & 1 & 244.2M & 3.4B   & 153   & 768 \\
        \midrule
        OGBL-COLLAB~\cite{hu2020open}    & Link & 1 & 235.9K & 1.3M      & - & 128 \\
        OGBL-PPA~\cite{szklarczyk2019string}        & Link & 1 & 576.3K & 30.3M     & - & 58 \\
        \midrule
        MUTAG~\cite{debnath1991structure}       & Graph & 188   & 17.9 & 57.5   & 2 & 7 \\
        PTC~\cite{toivonen2003statistical}         & Graph & 344   & 25.6 & 77.5   & 2 & 19 \\
        PROTEINS~\cite{borgwardt2005protein}    & Graph & 1113  & 39.1 & 184.7  & 2 & 3 \\
        COLLAB~\cite{yanardag2015structural}      & Graph & 5000  & 74.5 & 4989.5 & 3 & 367 \\
        \bottomrule
        
    \end{tabular}
    \label{tab:data_statistics}
\end{table}

\section{Evaluation}
\label{sec:evaluation}
%%CL \noindent\textbf{Configurations.} 
We run experiments on a multi-GPU server with four NVIDIA Geforce 1080Ti GPUs (11 GB memory for each), 16-core Intel Xeon CPU (2.10 GHz), and 512 GB of RAM. It runs PyTorch \cite{Pytorch} and DGL \cite{DGL}.
%%CL \noindent\textbf{GNN models, tasks, and datasets.} 
We train 5 GNN models such as GraphSAGE, GAT, and GIN to fulfill 3 kinds of GNN tasks, namely, node property prediction, link property prediction and graph property regression, over 9 graph datasets. For the heterogeneous graph MAG240M, we transform it into an undirected homogeneous graph and generate features for author and institution nodes by averaging their neighbors' features, which doubles its structure and feature information, in total about 400GB. Table \ref{tab:data_statistics} shows the statistics of the datasets.

%%CL \noindent\textbf{Deployed systems.} 
We use the full-precision training as the natural baseline, denoted by ``Full''. We also run the two variants of {\qmethod}, {\qmethod}-SQ denoted by ``SQ'' and {\qmethod}-VQ by ``VQ''. In addition, we explore the joint effects of enabling PaGraph's caching and quantization methods. We use optimal hyperparameter setup for the full precision baseline. Meanwhile, for each quantized training task, we use the same model setup as ``Full'' with minor changes in learning rate and dropout probability. 
 
%%CL \noindent\textbf{Metrics.} We first understand the negligible accuracy effects of {\qmethod} in Section~\ref{sect:eval:acc} and then the training acceleration in Section~\ref{sect:eval:trainacce}. Finally, we verify that {\qmethod} scales to large graph and conduct ablation studies such as training time breakdown and multi-GPU speeds in Section~\ref{sect:eval:sgg}. %%CL   section, we will firstly have some accuracy results to show quantization doesn't cause much accuracy degradation, then we will show how quantization speeds up GNN training. 

\begin{table}[!t]
\small
    \caption{Node classification accuracy. CR is compression ratio, while Acc is the test set accuracy.}
    \label{node_acc}
    \centering
    \begin{tabular}{ llcccc } 
        \toprule
        \multicolumn{2}{c}{} & \multicolumn{2}{c}{Reddit}   & \multicolumn{2}{c}{OGBN-Papers100M}  \\
        Model & Method & CR & Acc(\%) & CR & Acc(\%) \\
        \midrule
        \multirow{3}{*}{GraphSAGE\cite{hamilton2017inductive}}  & Full  & -     & 96.3 & -      & 66.1      \\
                                    & SQ    & 32    & 96.4 & 32     & 65.4      \\
                                    & VQ    & 228.6 & 96.1 & 46.5   & 65.4      \\        
        \midrule
        \multirow{3}{*}{GAT\cite{velivckovic2017graph}}        & Full  & -     & 95.3  & -     & 65.8      \\
                                    & SQ    & 32    & 95.4  & 32    & 64.9      \\
                                    & VQ    & 163.2 & 95.3  & 36.6  & 65.3      \\    
        \midrule
        \multirow{3}{*}{ClusterGCN\cite{chiang2019cluster}}     & Full  & -     & 96.0 & -      & 63.3   \\
                                        & SQ    & 32    & 96.0 & 32     &  62.8  \\
                                        & VQ    & 148   & 95.5 & 46.5   &  62.7  \\  
        \midrule
        \multirow{3}{*}{MLP\cite{taud2018multilayer}}    & Full & - & 72.6 & - & 46.9    \\
                                & SQ & 32 & 68.1 & 32 & 35.2    \\
                                & VQ & 51.2 & 65.1 & 39.4 & 38.1    \\
        % \midrule
        % \multirow{3}{*}{SIGN} & Full  & 1 & 96.4 & 1 & 65.8 & 1 & 66.0    \\
        %     & SQ & 16 & 96.5 & 16 & 65.2 & 16 & 65.3   \\
        %     & VQ & 136 & 96.3 & 25.6 & 65.3 & 38.4 & 64.9   \\        
        \bottomrule
    \end{tabular}
\end{table}

\subsection{Accuracy Validation}
\label{sect:eval:acc}
%%CL There are three major applications of GNNs, node property prediction, link property prediction and graph property regression. 
%%CL For each application, we choose several widely used models and test their performance with and without quantization on different graph datasets.

\noindent\textbf{Node Property Prediction.}  
To test the accuracy of the node property prediction task, we choose to train four GNN models,GraphSAGE, GAT, ClusterGCN, and MLP, over two datasets, Reddit and OGBN-Papers100M. Though Reddit is a small dataset, it has been widely used as a reference for accuracy validation. %%CL medium sized graph dataset and the last two are among the largest public graph dataset. We also chose three models, GraphSAGE, GAT and ClusterGCN, since they are typical, widely used models and include different aggregation and sampling methods, we also did tests on MLP for comparison. 

Table~\ref{node_acc} summarizes the accuracy results as well as the compression ratios achieved by \qmethod. We draw the following findings. First, across all test cases, SQ achieves a constant compression ratio of 32, as it quantizes each 32 float number of features into an 1-bit integer. Unlike SQ, however, VQ can achieve significantly higher compression ratios, up to 228.6, while delivering comparable or slightly better accuracy than SQ. Second, compared to the full precision baseline, for models except MLP, both SQ and VQ achieve almost the same accuracy numbers. This implies that shows GNNs usually reach comparable accuracy with aggressive compression. However, MLP observes the maximum accuracy degradation of around 9\%, which because of its more sensitive to the input error.

\begin{table}[t!]
\small
    \centering
    \caption{Link prediction accuracy}
    \label{link_acc}        
    \begin{tabular}{ llcccc } 
        \toprule
        \multicolumn{2}{c}{} & \multicolumn{2}{c}{OGBL-PPA}   & \multicolumn{2}{c}{OGBL-COLLAB}   \\
        Model & Method & CR & Hits@100(\%) & CR & Hits@50(\%) \\
        \midrule
        \multirow{3}{*}{GCN\cite{GCN}}       & Full  &  -    & 18.4 &   -   &   50.6      \\
                                    & SQ   &  32   & 18.5 &  32   &   49.0   \\
                                    & VQ  &  46.5 & 18.9 &  46.5 &   49.2        \\   
        \midrule
        \multirow{3}{*}{GraphSAGE}  & Full  &  -    & 16.4 &   -   &  49.4   \\
                                    & SQ    &  32   & 16.3 &   32  &  50.4    \\
                                    & VQ    &  46.5 & 16.5 &  46.5 &  49.3    \\       
        \bottomrule
    \end{tabular}
  \label{tab:link}
\end{table}

%\begin{table}[h]
%    \centering
%    \caption{link prediction accuracy}
%    \label{link_acc}        
%    \begin{tabular}{ llcccc } 
%        \toprule
%        \multicolumn{2}{c}{} & \multicolumn{2}{c}{OGBL-PPA}   & \multicolumn{2}{c}{OGBL-COLLAB}   \\
%        Model & Method & CR & Hits@100(\%) & CR & Hits@50(\%) \\
%        \midrule
%        \multirow{3}{*}{GCN}        & Full  &  -    & 18.42 &   -   &   50.58      \\
%                                    & SQ    &  32   & 18.54 &  32   &   49.03   \\
%                                    & VQ    &  46.5 & 18.87 &  46.5 &   49.23        \\   
%        \midrule
%        \multirow{3}{*}{GraphSAGE}  & Full  &  -    & 16.39 &   -   &  49.42   \\
%                                    & SQ    &  32   & 16.37 &   32  &  50.41    \\
%                                    & VQ    &  46.5 & 16.46 &  46.5 &  49.28    \\       
%        \bottomrule
%    \end{tabular}
%  \label{tab:link}
%\end{table}

\begin{table}[h]
    \centering
    \small
    \caption{Graph classification accuracy}
    \label{graph_acc}        
    \begin{tabular}{ llcccccccc } 
        \toprule
        \multicolumn{2}{c}{} & \multicolumn{2}{c}{PTC}   & \multicolumn{2}{c}{MUTAG} & \multicolumn{2}{c}{PROTEINS}  & \multicolumn{2}{c}{COLLAB} \\
        Model & Method & CR & Acc(\%) & CR & Acc(\%) & CR & Acc(\%) & CR & Acc(\%) \\
        \midrule
        \multirow{3}{*}{GCN}    & Full  &  -    & 61.3  &  -    & 85.6  &   -   & 74.4 &   -   & 82.9       \\
                                & VQ    &  152  & 63.4  &  112  & 87.3  &  136  & 74.0 & 1468  & 83.1          \\   
        \midrule
        \multirow{3}{*}{GIN\cite{GIN}}   & Full   &  -    & 64.1  &  -    & 87.7  &   -   & 73.6 &   -   & 81.8          \\
                                & VQ    &  152  & 65.1  &  112  & 87.8  &  136  & 73.6 & 1468  & 82.2        \\       
        \bottomrule
    \end{tabular}
  \label{tab:graph}
\end{table}

\noindent\textbf{Link Property Prediction.} 
Further, we run experiments with link prediction tasks, which are to predict properties of edges (i.e. pairs of nodes). Here, we use two graph datasets, namely, OGBL-PPA and OGBL-COLLAB. OGBL-PPA is a protein-protein association network, organized as an undirected, unweighted graph, with Nodes representing different proteins from 58 species, and edges indicating biologically associations between proteins.  OGBL-COLLAB is an author collaboration network, also organized as an undirected graph. We train GCN and GraphSAGE models over these datasets, and report the prediction results in Table~\ref{tab:link}. Similar to the node property prediction tasks, using both \qmethod-SQ and \qmethod-VQ can achieve comparable link prediction accuracy (Hits@100 in OGBL-PPA, Hits@50 in OGBL-COLLAB) as the full precision baseline. Again, here, \qmethod-VQ shows better compression ratio, which is a constant value 46.5, that is 45\% higher than \qmethod-SQ. %%CL , we find models, GCN and GraphSage, after using BiFeat-SQ and BiFeat-VQ are slight improved their privacy. With Compression ratio 32 and 46.5 in BiFeat-SQ and BiFeat-VQ, which reduced the communication significantly, OGBL-PPA achieves best accuracy through BiFeat-VQ, OGBL-PPA also gets best accuracy by using BiFeat-SQ, respectively. This is because our proposed method can keep the most valuable information and get rid of noise so that strengthen model efficiency.

\noindent\textbf{Graph Property Prediction.} 
As for the graph property prediction tasks, we train GCN and GIN with four datasets PTC, MUTAG, PROTEINS and COLLAB. The first three datasets are bio-informatics datasets, while the last one is a social network dataset. Note that the features of these datasets are one-hot encoded, i.e., each feature vector can be represented by a single integer, indicating the index of the only non-zero element in such vector. Clearly, when being applied to such features, our scalar quantization is lossless and behave exactly the same as the full precision baseline. Therefore, we omit its results from Table~\ref{tab:graph}.  

On the other hand, since the one-hot codes theoretically have a high lossless compression ratio, to test the accuracy impact of vector quantization while stressing \qmethod, we set a higher compression ratio than the lossless setting. Doing so forces \qmethod-VQ to lose some information during quantization. Table \ref{tab:graph} demonstrates that impressively, \qmethod-VQ can actually enhance the training accuracy. Based on our observations, vector quantization extracts the valuable information and filters out noises, thus improving model robustness.

\begin{table}[t!]
\begin{minipage}[b]{0.45\linewidth}
\centering
\small
\caption{Per-epoch time. NC and C stand for settings with caching disabled/enabled}
\begin{tabular}{ llcccc } 
        \toprule
        \multicolumn{2}{c}{} & \multicolumn{2}{c}{Reddit}   & \multicolumn{2}{c}{Papers100M}  \\
        Model & Method & NC & C & NC & C  \\
        \midrule
        \multirow{3}{*}{\specialcell{Graph-\\ SAGE}}  & Full  & 11.60 & 2.47  & 118.5  & 103.1   \\
                                    & SQ    & 3.54  & 2.96  & 66.4  & 64.4       \\
                                    & VQ    & 3.19  & 2.81  & 69.8  & 67.1       \\
        \midrule
        \multirow{3}{*}{GAT}        & Full  & 13.00  & 3.86   &  195.3  & 187.5 \\
                                    & SQ    & 5.28  & 4.49   &  115.3  & 112.2            \\
                                    & VQ    & 4.88  & 4.42   &  117.9  & 112.0        \\
        \midrule
        \multirow{3}{*}{\specialcell{Cluster-\\ GCN}} & Full  & 2.23  & 1.68   & 12.91 & 7.24      \\
                                    & SQ    & 1.77  & 1.73   & 7.55  & 7.30      \\
                                    & VQ    & 1.86  & 1.79   & 8.54  & 7.33       \\                                    
        \bottomrule
    \end{tabular}
    \label{acceleration}
\end{minipage}
\hfill
\begin{minipage}[b]{0.45\linewidth}
\centering
\small
    \caption{Accuracy and per-epoch time with MAG240M}
    \begin{tabular}{llccc}
     \toprule
        \multicolumn{4}{c}{} & \multicolumn{1}{c}{Epoch} \\
        Model & Method & CR & Acc(\%) &  time (s) \\
        \midrule
         \multirow{3}{*}{\specialcell{Graph-\\ SAGE}}     & Full  & -     & 68.6  & 470.9  \\
                                        & SQ    & 32    & 68.5  & 114.3  \\
                                        & VQ    & 46.5  & 67.7  & 112.2  \\
        \midrule
         \multirow{3}{*}{GAT}           & Full  & -     & 67.8  & 509.7  \\
                                        & SQ    & 32    & 67.7  & 169.8  \\
                                        & VQ    & 46.5  & 67.7  & 163.4  \\
        \midrule
         \multirow{3}{*}{MLP}   & Full  & -     & 52.1  & 9.38 \\
                                & SQ    & 32    & 42.0  & 7.07 \\
                                & VQ    & 34.9  & 44.1  & 8.62 \\
        \bottomrule
    \end{tabular}
    \label{tab:giant_graph}
\end{minipage}
\end{table}

\subsection{Training Acceleration}
\label{sect:eval:trainacce}
 %%CL Mini-batch training method requires constantly loading sampled subgraphs into GPU, which slows down the whole training due to the large size of features of the subgraphs.
%%CL Compressing features significantly reduces amount of data needed to be loaded, thus data loading workload is reduced and it is no longer the bottleneck of training. 

Next, we shift our attention from accuracy to the training performance improvement. To this end, 
% We test training GraphSAGE,GAT,... on Reddit, OGBN-Products and MAG240M datasets. 
we measure the average running time per epoch of node property prediction tasks, where we train three GNN models with both Reddit and OGBN-Papers100M. Here, we additionally compare to another baseline, corresponding to PaGraph's caching, and also explore its joint effects with quantization. In Table~\ref{acceleration}, with no caching (NC), the two {\qmethod} quantization variants achieve greatest speedups at 3.27$\times$ (SQ) and 3.63$\times$ (VQ), for the three models, respectively. %%CL -3$\times$ speedup for GraphSAGE and GAT, . 
When additionally enabling cache, for Reddit, no visible speedups are observed. This is because Reddit can be fully cached into GPU memory, and the two optimizations, caching or quantization, both can completely eliminating data loading. Concerning the Papers100M dataset,
two models except ClusterGCN observes speedups of 1.53-1.64$\times$. This is because using cache in full precision features only gets a low speedup due to the proportion of cached node is too low. Feature compression can increase the cache rate and get a high hit rate, allows a higher speedup. Unlike this, combining caching with quantization does not improve performance of the full precision training with caching enabled. This is because the training set of ClusterGCN is relatively small and can be fully cached within GPU memory.

 %\begin{figure}[h]
 %    \centering
 %    \includegraphics[width=0.7\textwidth]{figs/speedup_breakdown.pdf}
 %    \caption{Detailed run time breakdown of GraphSAGE training on MAG240M with different quantization settings}
 %    \label{fig:speedup_breakdown}
 %\end{figure}

\begin{figure*}[t!]
\begin{minipage}[b]{\linewidth}
\centering
\subfigure[Loss \& Accuracy]{
\raggedleft
\includegraphics[width=0.47\textwidth]{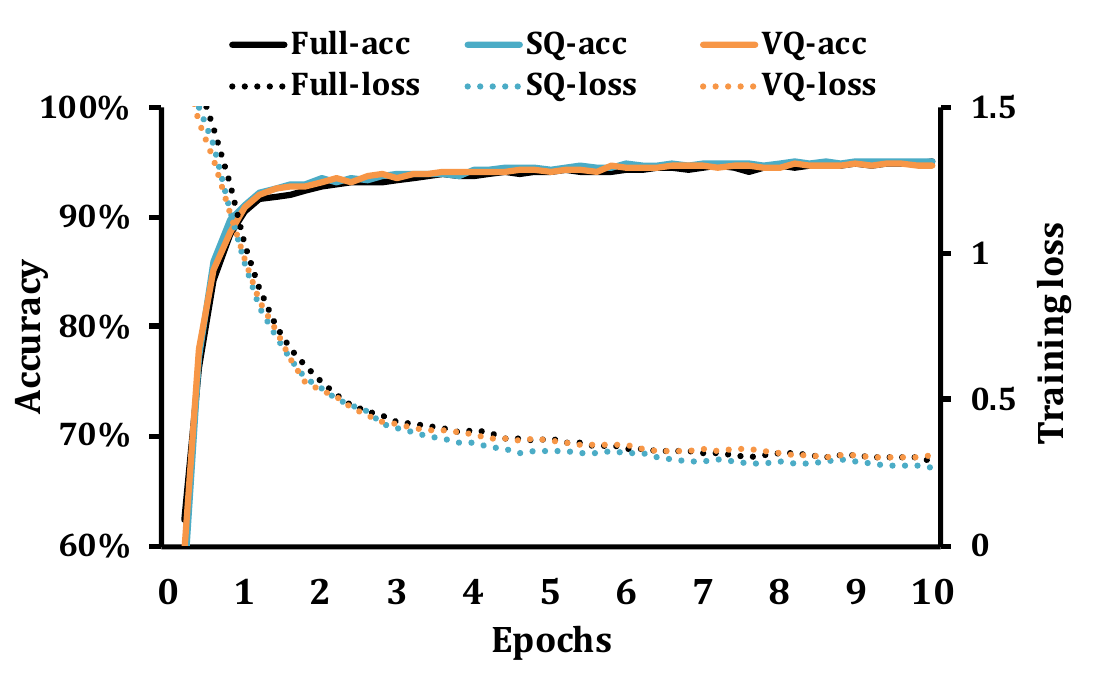}
\label{fig:convergence_rate}
}
\hfill
\subfigure[Breakdown]{
\raggedleft
\includegraphics[width=0.24\linewidth]{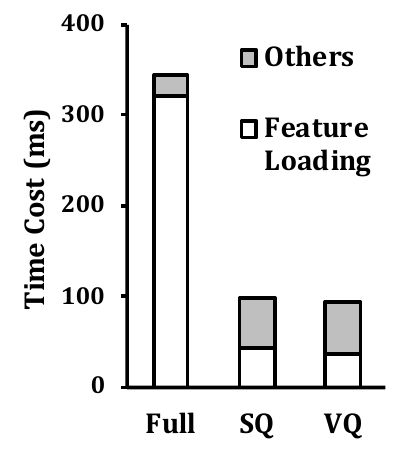}
\label{fig:contention-create}
}%
\hfill
\subfigure[Multi-GPU]{ 
\raggedleft
\includegraphics[width=0.24\linewidth]{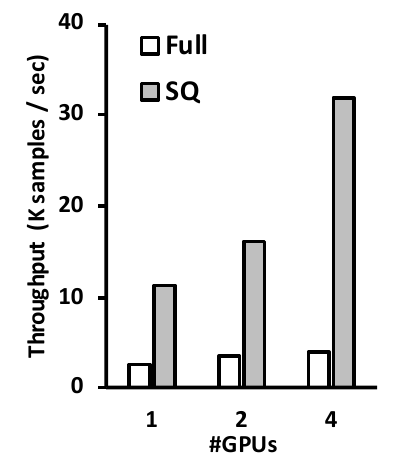}
\label{fig:contention-mkdir}
}%
\captionof{figure}{Loss and test set accuracy of GraphSAGE training on Reddit with different quantization settings (a); Detailed run time breakdown of GraphSAGE training on MAG240M with different quantization settings (b); Multi-GPU acceleration with GraphSAGE and MAG240M dataset (c)}
\end{minipage}
\end{figure*}

 In addition to the per-epoch speedup analysis, we also investigate  how the convergence rate evolves as the training runs from epoch to epoch. Here, due to space limit, we choose to train GraphSAGE over Reddit with the full precision baseline and two {\qmethod} variants, and report the test accuracy results in Figure~\ref{fig:convergence_rate}. The conclusion drawn here is that using \qmethod, regardless of using SQ or VQ, the quantization-based training can converge to almost the same accuracy without introducing more epochs than the full precision baseline. Together with the above per-epoch performance speedup, we can conclude that \qmethod is able to accelerate the end-to-end training performance of GNNs.
% \begin{figure}[h]
%     \centering
%     \includegraphics[width=0.7\textwidth]{figs/convergence_rate.pdf}
%     \caption{Loss and test set accuracy of GraphSAGE training on Reddit with different quantization settings}
%     \label{fig:convergence_rate}
% \end{figure}

\subsection{Supporting Giant Graph}
\label{sect:eval:sgg}

Finally, we validate if \qmethod can accelerate the training of extremely large graphs. Here, we choose MAG240M, the largest graph among our datasets, and run the node property prediction tasks with three models, namely, GraphSAGE, GAT and MLP. We first report the accuracy and per-epoch time results in Table~\ref{tab:giant_graph}. For GraphSAGE and GAT, the two \qmethod variants achieve comparable accuracy as the full precision baseline. On the per-epoch time aspect, \qmethod-SQ and \qmethod-VQ introduce speedups of 3.00-4.12x$\times$ and 3.12-4.20$\times$ for the two models, respectively. However, similar to the above results, MLP observes the least benefits of \qmethod, for instance, \qmethod  introduces a maximum accuracy loss of 10\%, while only introducing speedups of up to 1.33$\times$. %%CL First Table, Second Reduction, Finally, multi-GPU.

Second, we conduct the ablation studies to understand the reasons for {\qmethod} to accelerate the training performance, especially with large graphs. To do so, we measure the changes of the ratio of feature loading across different settings, and summarize the results in Figure~\ref{fig:contention-create}. Within the full precision training, the feature loading is the bottleneck. However, when adopting either SQ or VQ, the feature load time cost has been reduced by up to 90\%. We also observe slight increases in the time cost of other tasks, mainly due to the extra dequantization overhead. %%CL   reduction of data amount needed to be copied to GPU, as  shows, the data collection and loading phase takes most of the run time, and compression lowers both of them.

Finally, to complement the above single-GPU experiments, here, we run multi-GPU training with MAG240M. As depicted in Figure~\ref{fig:contention-mkdir}, without quantization, adding more GPUs does not necessarily bring performance improvements. This is because the resource contention among training jobs over multiple GPUs exacerbate the feature loading bottleneck. In contrast, using quantization, training performance scales well, due to the reduction in CPU usage and PCIe bandwidth consumption. In the end, we manage to train GraphSAGE over the largest MAG240M dataset using 4 GPUs within one hour, introducing a speedup of up to 8.6$\times$ comparing to the full precision baseline, which has to run over 7.5 hours.

\section{Conclusion}
We pioneer the adoption of the feature quantization methods to break the GPU memory limit and eliminate the feature loading bottleneck, faced by GNN mini-batch training over large graphs. Following this, we materialize our ideas into {\qmethod}, a GNN feature quantization methodology that supports both Scalar and Vector quantization approaches. Furthermore, we provide theoretical proofs and run extensive experiments to validate that {\qmethod} can greatly reduce the feature data loading volumes and shorten the end-to-end training time, with bounded accuracy loss.

 %%CL (we give a analysis of upbound of this accuracy degradation on most popular GNN models.)  With feature compression and cache, it is possible to efficiently train large scale graphs with a few machines. Our method is easy to implement, only a few lines of code need to be added.

\clearpage
\bibliographystyle{plain}
\bibliography{references}

\begin{thebibliography}{10}

\bibitem{DGL}
{DGL Homepage}.
\newblock \url{https://www.dgl.ai/}.
\newblock [Online; accessed May-2022].

\bibitem{Pytorch}
{Pytorch Homepage}.
\newblock \url{https://pytorch.org/}.
\newblock [Online; accessed May-2022].

\bibitem{aji2017sparse}
Alham~Fikri Aji and Kenneth Heafield.
\newblock Sparse communication for distributed gradient descent.
\newblock {\em arXiv preprint arXiv:1704.05021}, 2017.

\bibitem{allen2019convergence}
Zeyuan Allen-Zhu, Yuanzhi Li, and Zhao Song.
\newblock A convergence theory for deep learning via over-parameterization.
\newblock In {\em International Conference on Machine Learning}, pages
  242--252. PMLR, 2019.

\bibitem{bahri2021binary}
Mehdi Bahri, Ga{\'e}tan Bahl, and Stefanos Zafeiriou.
\newblock Binary graph neural networks.
\newblock In {\em Proceedings of the IEEE/CVF Conference on Computer Vision and
  Pattern Recognition}, pages 9492--9501, 2021.

\bibitem{bojchevski2017deep}
Aleksandar Bojchevski and Stephan G{\"u}nnemann.
\newblock Deep gaussian embedding of graphs: Unsupervised inductive learning
  via ranking.
\newblock {\em arXiv preprint arXiv:1707.03815}, 2017.

\bibitem{borgwardt2005protein}
Karsten~M Borgwardt, Cheng~Soon Ong, Stefan Sch{\"o}nauer, SVN Vishwanathan,
  Alex~J Smola, and Hans-Peter Kriegel.
\newblock Protein function prediction via graph kernels.
\newblock {\em Bioinformatics}, 21(suppl\_1):i47--i56, 2005.

\bibitem{chen2018fastgcn}
Jie Chen, Tengfei Ma, and Cao Xiao.
\newblock Fastgcn: fast learning with graph convolutional networks via
  importance sampling.
\newblock {\em arXiv preprint arXiv:1801.10247}, 2018.

\bibitem{chen2020simple}
Ming Chen, Zhewei Wei, Zengfeng Huang, Bolin Ding, and Yaliang Li.
\newblock Simple and deep graph convolutional networks.
\newblock In {\em International Conference on Machine Learning}, pages
  1725--1735. PMLR, 2020.

\bibitem{chereda2021explaining}
Hryhorii Chereda, Annalen Bleckmann, Kerstin Menck, J{\'u}lia Perera-Bel,
  Philip Stegmaier, Florian Auer, Frank Kramer, Andreas Leha, and Tim
  Bei{\ss}barth.
\newblock Explaining decisions of graph convolutional neural networks:
  patient-specific molecular subnetworks responsible for metastasis prediction
  in breast cancer.
\newblock {\em Genome medicine}, 13(1):1--16, 2021.

\bibitem{chiang2019cluster}
Wei-Lin Chiang, Xuanqing Liu, Si~Si, Yang Li, Samy Bengio, and Cho-Jui Hsieh.
\newblock Cluster-gcn: An efficient algorithm for training deep and large graph
  convolutional networks.
\newblock In {\em Proceedings of the 25th ACM SIGKDD International Conference
  on Knowledge Discovery \& Data Mining}, pages 257--266, 2019.

\bibitem{debnath1991structure}
Asim~Kumar Debnath, Rosa~L Lopez~de Compadre, Gargi Debnath, Alan~J Shusterman,
  and Corwin Hansch.
\newblock Structure-activity relationship of mutagenic aromatic and
  heteroaromatic nitro compounds. correlation with molecular orbital energies
  and hydrophobicity.
\newblock {\em Journal of medicinal chemistry}, 34(2):786--797, 1991.

\bibitem{devlin2018bert}
Jacob Devlin, Ming-Wei Chang, Kenton Lee, and Kristina Toutanova.
\newblock Bert: Pre-training of deep bidirectional transformers for language
  understanding.
\newblock {\em arXiv preprint arXiv:1810.04805}, 2018.

\bibitem{ding2021vq}
Mucong Ding, Kezhi Kong, Jingling Li, Chen Zhu, John Dickerson, Furong Huang,
  and Tom Goldstein.
\newblock Vq-gnn: A universal framework to scale up graph neural networks using
  vector quantization.
\newblock {\em Advances in Neural Information Processing Systems}, 34, 2021.

\bibitem{dong2021global}
Jialin Dong, Da~Zheng, Lin~F Yang, and Geroge Karypis.
\newblock Global neighbor sampling for mixed cpu-gpu training on giant graphs.
\newblock {\em arXiv preprint arXiv:2106.06150}, 2021.

\bibitem{feng2020sgquant}
Boyuan Feng, Yuke Wang, Xu~Li, Shu Yang, Xueqiao Peng, and Yufei Ding.
\newblock Sgquant: Squeezing the last bit on graph neural networks with
  specialized quantization.
\newblock In {\em 2020 IEEE 32nd International Conference on Tools with
  Artificial Intelligence (ICTAI)}, pages 1044--1052. IEEE, 2020.

\bibitem{fey2021gnnautoscale}
Matthias Fey, Jan~E Lenssen, Frank Weichert, and Jure Leskovec.
\newblock Gnnautoscale: Scalable and expressive graph neural networks via
  historical embeddings.
\newblock In {\em International Conference on Machine Learning}, pages
  3294--3304. PMLR, 2021.

\bibitem{fey2019fast}
Matthias Fey and Jan~Eric Lenssen.
\newblock Fast graph representation learning with pytorch geometric.
\newblock {\em arXiv preprint arXiv:1903.02428}, 2019.

\bibitem{gao2019convergence}
Ruiqi Gao, Tianle Cai, Haochuan Li, Cho-Jui Hsieh, Liwei Wang, and Jason~D Lee.
\newblock Convergence of adversarial training in overparametrized neural
  networks.
\newblock {\em Advances in Neural Information Processing Systems}, 32, 2019.

\bibitem{gersho2012vector}
Allen Gersho and Robert~M Gray.
\newblock {\em Vector quantization and signal compression}, volume 159.
\newblock Springer Science \& Business Media, 2012.

\bibitem{gray1984vector}
Robert Gray.
\newblock Vector quantization.
\newblock {\em IEEE Assp Magazine}, 1(2):4--29, 1984.

\bibitem{haase2020dependent}
Paul Haase, Heiko Schwarz, Heiner Kirchhoffer, Simon Wiedemann, Talmaj Marinc,
  Arturo Marban, Karsten M{\"u}ller, Wojciech Samek, Detlev Marpe, and Thomas
  Wiegand.
\newblock Dependent scalar quantization for neural network compression.
\newblock In {\em 2020 IEEE International Conference on Image Processing
  (ICIP)}, pages 36--40. IEEE, 2020.

\bibitem{hamilton2017inductive}
Will Hamilton, Zhitao Ying, and Jure Leskovec.
\newblock Inductive representation learning on large graphs.
\newblock {\em Advances in neural information processing systems}, 30, 2017.

\bibitem{GCN}
Negar Heidari and Alexandros Iosifidis.
\newblock Progressive graph convolutional networks for semi-supervised node
  classification.
\newblock {\em IEEE Access}, 9:81957--81968, 2021.

\bibitem{heinly2012comparative}
Jared Heinly, Enrique Dunn, and Jan-Michael Frahm.
\newblock Comparative evaluation of binary features.
\newblock In {\em European Conference on Computer Vision}, pages 759--773.
  Springer, 2012.

\bibitem{hu2020open}
Weihua Hu, Matthias Fey, Marinka Zitnik, Yuxiao Dong, Hongyu Ren, Bowen Liu,
  Michele Catasta, and Jure Leskovec.
\newblock Open graph benchmark: Datasets for machine learning on graphs.
\newblock {\em Advances in neural information processing systems},
  33:22118--22133, 2020.

\bibitem{hu2020gpt}
Ziniu Hu, Yuxiao Dong, Kuansan Wang, Kai-Wei Chang, and Yizhou Sun.
\newblock Gpt-gnn: Generative pre-training of graph neural networks.
\newblock In {\em Proceedings of the 26th ACM SIGKDD International Conference
  on Knowledge Discovery \& Data Mining}, pages 1857--1867, 2020.

\bibitem{hu2016BNN}
Itay Hubara, Matthieu Courbariaux, Daniel Soudry, Ran El-Yaniv, and Yoshua
  Bengio.
\newblock Binarized neural networks.
\newblock In {\em Proceedings of the 30th International Conference on Neural
  Information Processing Systems}, NIPS'16, page 4114–4122, Red Hook, NY,
  USA, 2016. Curran Associates Inc.

\bibitem{kipf2016semi}
Thomas~N Kipf and Max Welling.
\newblock Semi-supervised classification with graph convolutional networks.
\newblock {\em arXiv preprint arXiv:1609.02907}, 2016.

\bibitem{lin2017dgc}
Yujun Lin, Song Han, Huizi Mao, Yu~Wang, and William~J Dally.
\newblock Deep gradient compression: Reducing the communication bandwidth for
  distributed training.
\newblock {\em arXiv preprint arXiv:1712.01887}, 2017.

\bibitem{lin2020pagraph}
Zhiqi Lin, Cheng Li, Youshan Miao, Yunxin Liu, and Yinlong Xu.
\newblock Pagraph: Scaling gnn training on large graphs via computation-aware
  caching.
\newblock In {\em Proceedings of the 11th ACM Symposium on Cloud Computing},
  pages 401--415, 2020.

\bibitem{liu2020towards}
Meng Liu, Hongyang Gao, and Shuiwang Ji.
\newblock Towards deeper graph neural networks.
\newblock In {\em Proceedings of the 26th ACM SIGKDD international conference
  on knowledge discovery \& data mining}, pages 338--348, 2020.

\bibitem{liu2020BRN}
Zechun Liu, Wenhan Luo, Baoyuan Wu, Xin Yang, Wei Liu, and Kwang-Ting Cheng.
\newblock Bi-real net: Binarizing deep network towards real-network
  performance.
\newblock {\em Int. J. Comput. Vision}, 128(1):202–219, jan 2020.

\bibitem{rodriguez2018beyond}
Pau Rodr{\'\i}guez, Miguel~A Bautista, Jordi Gonz{\`a}lez, and Sergio Escalera.
\newblock Beyond one-hot encoding: Lower dimensional target embedding.
\newblock {\em Image and Vision Computing}, 75:21--31, 2018.

\bibitem{seide20141onebit}
Frank Seide, Hao Fu, Jasha Droppo, Gang Li, and Dong Yu.
\newblock 1-bit stochastic gradient descent and its application to
  data-parallel distributed training of speech dnns.
\newblock In {\em Fifteenth Annual Conference of the International Speech
  Communication Association}, 2014.

\bibitem{stone2010gpu}
John~E Stone, David~J Hardy, Ivan~S Ufimtsev, and Klaus Schulten.
\newblock Gpu-accelerated molecular modeling coming of age.
\newblock {\em Journal of Molecular Graphics and Modelling}, 29(2):116--125,
  2010.

\bibitem{strom2015TBQ}
Nikko Strom.
\newblock Scalable distributed dnn training using commodity gpu cloud
  computing.
\newblock In {\em Proceedings of Sixteenth Annual Conference of the
  International Speech Communication Association}, 2015.

\bibitem{szklarczyk2019string}
Damian Szklarczyk, Annika~L Gable, David Lyon, Alexander Junge, Stefan Wyder,
  Jaime Huerta-Cepas, Milan Simonovic, Nadezhda~T Doncheva, John~H Morris, Peer
  Bork, et~al.
\newblock String v11: protein--protein association networks with increased
  coverage, supporting functional discovery in genome-wide experimental
  datasets.
\newblock {\em Nucleic acids research}, 47(D1):D607--D613, 2019.

\bibitem{tailor2020degree}
Shyam~A Tailor, Javier Fernandez-Marques, and Nicholas~D Lane.
\newblock Degree-quant: Quantization-aware training for graph neural networks.
\newblock {\em arXiv preprint arXiv:2008.05000}, 2020.

\bibitem{taud2018multilayer}
Hind Taud and JF~Mas.
\newblock Multilayer perceptron (mlp).
\newblock In {\em Geomatic approaches for modeling land change scenarios},
  pages 451--455. Springer, 2018.

\bibitem{toivonen2003statistical}
Hannu Toivonen, Ashwin Srinivasan, Ross~D King, Stefan Kramer, and Christoph
  Helma.
\newblock Statistical evaluation of the predictive toxicology challenge
  2000--2001.
\newblock {\em Bioinformatics}, 19(10):1183--1193, 2003.

\bibitem{velivckovic2017graph}
Petar Veli{\v{c}}kovi{\'c}, Guillem Cucurull, Arantxa Casanova, Adriana Romero,
  Pietro Lio, and Yoshua Bengio.
\newblock Graph attention networks.
\newblock {\em arXiv preprint arXiv:1710.10903}, 2017.

\bibitem{wang2019deep}
Minjie Wang, Da~Zheng, Zihao Ye, Quan Gan, Mufei Li, Xiang Song, Jinjing Zhou,
  Chao Ma, Lingfan Yu, Yu~Gai, et~al.
\newblock Deep graph library: A graph-centric, highly-performant package for
  graph neural networks.
\newblock {\em arXiv preprint arXiv:1909.01315}, 2019.

\bibitem{wen2017terngrad}
Wei Wen, Cong Xu, Feng Yan, Chunpeng Wu, Yandan Wang, Yiran Chen, and Hai Li.
\newblock Terngrad: Ternary gradients to reduce communication in distributed
  deep learning.
\newblock In {\em Proceedings of Advances in neural information processing
  systems}, pages 1509--1519, 2017.

\bibitem{wu2020graph}
Yongji Wu, Defu Lian, Yiheng Xu, Le~Wu, and Enhong Chen.
\newblock Graph convolutional networks with markov random field reasoning for
  social spammer detection.
\newblock In {\em Proceedings of the AAAI Conference on Artificial
  Intelligence}, volume~34, pages 1054--1061, 2020.

\bibitem{GIN}
Keyulu Xu, Weihua Hu, Jure Leskovec, and Stefanie Jegelka.
\newblock How powerful are graph neural networks?
\newblock In {\em International Conference on Learning Representations}, 2019.

\bibitem{xu2019cross}
Kun Xu, Liwei Wang, Mo~Yu, Yansong Feng, Yan Song, Zhiguo Wang, and Dong Yu.
\newblock Cross-lingual knowledge graph alignment via graph matching neural
  network.
\newblock {\em arXiv preprint arXiv:1905.11605}, 2019.

\bibitem{yanardag2015structural}
Pinar Yanardag and SVN Vishwanathan.
\newblock A structural smoothing framework for robust graph comparison.
\newblock {\em Advances in neural information processing systems}, 28, 2015.

\end{thebibliography}
\medskip{}
\clearpage

\appendix

\section{Proof of Theorem \ref{convergence}}
\label{sec:proof}

\begin{table}[!t]
\small
    \caption{Summary of notations.}
    \label{notations}
    \centering
    \begin{tabular}{ l p{10cm}} 
        \toprule
        Notation & Description  \\
        \midrule
        $n$             & Number of nodes in the graph. \\
        $d$             & Input features dimension.\\       
        $m$             & Hidden layer dimension, also known as the width of network.\\  
        $N(i)$          & Set of nodes adjacent to $i$.\\
        $L$             & Number of layers.\\   
        $[n]$           & $\{1,2, \cdot\cdot\cdot, n\}$.\\
        $\boldsymbol{X}$    & Input features of all nodes, with shape $n \times d$. Note that we also use $x_i$ to denote features of node $i$.\\
        $\boldsymbol{X}'$   & Quantized input features of all nodes with the same shape as $\boldsymbol{X}$.\\
        $\boldsymbol{A}$    & Input layer weights with the shape of $m\times d$.\\
        $\boldsymbol{W}$    & Hidden layer weights $\boldsymbol{W} = (\boldsymbol{W}_1, \cdot\cdot\cdot, \boldsymbol{W}_L)$, where weights of each layer has the shape of $m\times d$.\\
        $a$             & Output layer weights with the shape of $m\times 1$.\\
        $\boldsymbol{D}$    & Diagonal matrix, $\boldsymbol{D}_{i,l} = diag(I(\overline{g}_{i,l}\ge 0))$, where $I$ is the element-wise indicator function.\\        
        $f_i(\boldsymbol{W},\boldsymbol{X})$        & Output of node $i$ with weights $\boldsymbol{W}$ and input $\boldsymbol{X}$\\
        $l$             & Loss function\\  
        $Loss(\boldsymbol{W},\boldsymbol{X})$        & Loss with weights $\boldsymbol{W}$ and inputs $\boldsymbol{X}$\\
        \midrule
        $B(R)$          & The neighborhood of initial weights, $B(R) = \{\boldsymbol{W}:\|\boldsymbol{W}_l-\boldsymbol{W}^{(0)}_l\|_F\le R/\sqrt{m}, l\in [L]\}$\\
        $\mathcal{P}_{B(R)}$    & Euclidean projection to the convex set $B(R)$.\\
        $\alpha$        & Step size of projected gradient descent.\\
        $T$             & Number of steps.\\        
        \midrule
        $C^f_{i,l}$     & The expected scale of node $i$'s features after the aggregation of $l$ layers.\\
        $C^e_{i,l}$     & The expected scale of node $i$'s error caused by quantization after the aggregation of $l$ layers.\\
        $\overline{C}^f_{L}$    & The mean value of $C^f_{i,L}$.\\    
        $\overline{C}^e_{L}$    & The mean value of $C^e_{i,L}$.\\
        $\hat{C}_L$     & $C^e_{i,L} / C^f_{i,L}$.\\
        \bottomrule
    \end{tabular}
\end{table}

We organize the proof as follows. First, we formally define the structure of graph neural network (GNN) in Section~\ref{definition}. Then we give the detailed proof throughout Section~\ref{bound_smooth}-~\ref{final_proof}: (1) we prove that the gradient is bounded and the loss function is almost convex within the neighborhood of random initialization in Section~\ref{bound_smooth}; (2) we prove that with the optimal weights of uncompressed features, quantized features produce a bounded loss compared to the uncompressed features in Section~\ref{similar_loss}; and (3) with the help of the lemmas introduced in (1) and (2), we prove Theorem~\ref{convergence} in Section~\ref{final_proof}.
Finally, we present an analysis of the factors that may affect the compression ratio in Section~\ref{factor_analysis}.

\subsection{GNN Definitions}
\label{definition}

As Graph Convolutional Network (GCN) is one of the most widely-used GNN models, we use it to drive our theoretical proofs and analysis. Here, we follow the literature~\cite{kipf2016semi} to define the GCN structure as follows: %%CL \cheng{Why only based on GCN? Can it be generalized to other GNNs?}\yuxin{GCN is the backbone of most GNNs and different sampling methods only change the value of $\hat{C}$ }

$$
\begin{array}{l} 
g_{i,0} = \boldsymbol{A}x_{i} \\
h_{i,0} = \phi(g_{i,0}) \\
g_{i,l} = \boldsymbol{W}_lh_{i,l-1} \quad\quad\quad\  , for \ i \in [n] \ and \ l \in [L]\\
\overline{g}_{i,l} = \sum_{j\in N(i)} \frac{1}{|N(i)|}g_{j,l}\quad , for \   i \in [n] \   and \   l \in [L]\\
% \overline{g}_{i,l} = G_lL_{:,i}\quad , for \   i \in [n] \   and \   l \in [L]\\
h_{i,l} = \phi(\overline{g}_{i,l}) \quad\quad\quad\quad\ \, , for \   i \in [n] \   and \   l \in [L]\\
f_{i}(W,X) = a^{\top}h_{i,L}
\end{array}
$$

Here $x_i$ is the input feature, $g_{i,l}$ and $h_{i,l}$ are the feature vectors of node $i$ before and after the ReLU activation function $\phi$ of layer $l$. Note that these notations and definitions %%CL \yanc{could you be specific what is above?} 
are similar with those traditional DNN architecture studied in \cite{gao2019convergence, allen2019convergence}. 
%%CL \yanc{it says following GCN above, but then it is the same as DNN?}
However, the main difference lies in that GNN introduces an aggregation phase i.e. $\overline{g}_{i,l}$, where it gathers information by averaging neighbor features. It can also be written in a GCN style as $ \overline{g}_{i,l} = 
\boldsymbol{G}_l\boldsymbol{L}_{:,i}$, where $\boldsymbol{G}_l = (g_{1,l}, \cdot\cdot\cdot, g_{n,l})$.
We set $\boldsymbol{L}=\hat{\boldsymbol{D}}^{-1}\hat{\boldsymbol{A}}$,   where $\boldsymbol{\hat{A}}$ and $\boldsymbol{\hat{D}}$ are the adjacency matrix and degree matrix including self loop. 
We use the diagonal matrix $\boldsymbol{D}$ to indicate the consequences of ReLU function, so $h_{i,l}$ can be written as $\boldsymbol{D}_{i,l}\overline{g}_{i,l}$.

With $\boldsymbol{S} = \{x\in \mathbb{R}^{d}:\|x\|_{2}=1\}$, we scale the input features to $\boldsymbol{X}=\{x_{i} \in \boldsymbol{S} : i \in [n]\}$ following~\cite{allen2019convergence}.
We also follow the same initialization method as Definition 2.3 in~\cite{allen2019convergence} with a minor change on the output layer: $a_j \sim \mathcal{N}(0,1/\overline{C}^f_L)$. This is intended to ensure the output scale fits the label, $\overline{C}^f_L$ is a factor we discuss in Section~\ref{factor_analysis}. %%CL\yanc{be specific where.} 
We only train hidden layer weights $\boldsymbol{W}$ while keeping input and output layer weights $\boldsymbol{A}$ and $a$ with random initialization as \cite{allen2019convergence} does.

We denote $\mathcal{P}_{B(R)}$ as the Euclidean projection to the convex set $B(R)$ and $\boldsymbol{W}^{(t)}$ as the Weights after $t$ iterations.
We run projected gradient descent based on the constraint set $B(R)$ with step size $\alpha$ for $T$ steps using the following update rule.
$$
\begin{array}{l} 
\boldsymbol{V}^{(t+1)} = \boldsymbol{W}^{(t)} - \alpha \nabla_{\boldsymbol{W}} Loss(\boldsymbol{W}^{(t)}, \boldsymbol{X}),\\
\boldsymbol{W}^{(t+1)} = \mathcal{P}_{B(R)}(\boldsymbol{V}^{(t+1)})
\end{array}
$$

To measure the effect of aggregation, we introduce two factors $C^{e}_{i,l}$ and $C^{f}_{i,l}$ to represent the expectation of scaling of $h_{i,l}$ caused by the aggregation of $l$ layers. These factors have the following properties:

$$C^{e}_{i,0} = 1, \quad  C^{f}_{i,0}=1$$
$$C^{e}_{i,l+1} = \sqrt{\sum_{j\in N(i)}\sum_{k\in N(i)}r^e_{l,j,k}\frac{1}{|N(i)|^2}C^{e}_{j,l}C^{e}_{k,l}}, \quad C^{f}_{i,l+1} = \sqrt{\sum_{j\in N(i)}\sum_{k\in N(i)}r^f_{l,j,k}\frac{1}{|N(i)|^2}C^{f}_{j,l}C^{f}_{k,l}}$$
$$ C^{e}_{i,l} \le 1, \quad C^{f}_{i,l} \le 1 $$
% due to $r_{j,j}=1$, $r_{j,k}\le1$and with high probability
% $$0 \le r^e_{j,k} \le r^f{j,k}$$
% $$ C^{e}_{i,l} \le C^{f}_{i,l} $$
% for $r_{j,k}\ge r_{i,j}r_{i,k} \ge \omega^2 \ge 0 $ 
where $r^f_{l,j,k}$ and $r^e_{l,j,k}$ denotes the correlation of feature components and errors in layer $l$ between the node $j$ and node $k$, %%CL \cheng{r is not in the notation table} \yuxin{it is a notation only used locally here, to introduce some properties of C, we will discuss C layer without r, so i don't think it's worth put in the table}
respectively. 
With Assumption~\ref{iid_error}, $r^e_{0,j,k} = 0$ ($j\neq k$), $C^{e}_{i,l}$ is determined by the graph structure. $C^{f}_{i,l}$ is also relative to the graph structure, but it is also affected by the correlation of node features. 
$ C^{e}_{i,l} \le C^{f}_{i,l} $ has a high probability as $r^f_{0,j,k} > 0$ has high probability. So we have $ \frac{C^{e}_{i,l}}{C^{f}_{i,l}} \le 1$.
$C^{e}_{i,l}$ and $C^{f}_{i,l}$ can also be seen as weighted average of $l$-hop neighbors, thus we have $C^{e}_{i,l}\ge \frac{1}{n}$ and $C^{f}_{i,l} \ge \frac{1}{n}$. 
Therefore, we expect the value of $C^{e}_{i,l}$ , $C^{f}_{i,l}$, and $C^{e}_{i,l}$ / $C^{f}_{i,l}$ gets smaller as the layer $l$ gets deeper.
% The averaging would reduce the scale of error since they are i.i.d., but the overlap of $l$-hop neighbors makes some weights far bigger than others, weakens the reduction, so the factor $C^{e}_{i,l}$ is relative to both degrees and the overlap of $l$-hop neighbors. 
While the exact value of these factors could not be directly calculated with the statistics of graph, we can obtain them through quick profiling tests and derive the bound from the analysis. We conduct an analysis of these factors in Section~\ref{factor_analysis}. %%CL \yanc{please be specific which section.}

\subsection{Gradient Bounds and Almost Convex}
\label{bound_smooth}
In this subsection, we will prove that the gradient is bounded and the loss function is almost convex with regard to weights within the neighborhood of random initialization. We use the same proof sketch as \cite{gao2019convergence}, but our proof differs by having fixed input features.

%%CL \yanc{need a brief summary what this section is about.}

\newtheorem{lemma1}[theorem]{Lemma}
\begin{lemma1}
\label{A1}
If $ m \ge d $, with probability $1-O(nL)e^{-\Omega(m)}$ at initialization, we have $\|\boldsymbol{A}\|_{2}=O(1)$, $\|\boldsymbol{W}_L\|_{2}=O(1), \forall l \in [L]$, and $\|a\|_{2}=O(\frac{\sqrt{m}}{\overline{C}^f_L})$.

\end{lemma1}

\newtheorem{lemma2}[theorem]{Lemma}
\begin{lemma2}
\label{A2}
For any fixed input $\boldsymbol{X}=\{x_{i} \in \boldsymbol{S}: i \in [n]\}$, with probability $1-O(L)e^{-\Omega(m/L)}$ over the randomness of initialization, we have $\forall i\in [n]$ and $ l\in \{0,...,L\}$, 
% $$\|g_{i,l}\|_{2} \in[\frac{2\sqrt{2}}{3}C^{f}_{i,l-1}, \frac{4\sqrt{2}}{3}C^{f}_{i,l-1}], \quad \|g'_{i,l}-g_{i,l}\|_{2} \in[\frac{2\sqrt{2}}{3}\delta C^{e}_{i,l-1}, \frac{4\sqrt{2}}{3}\delta C^{e}_{i,l-1}]$$
% $$\|\overline{g}_{i,l}\|_{2} \in[\frac{2\sqrt{2}}{3}C^{f}_{i,l}, \frac{4\sqrt{2}}{3}C^{f}_{i,l}], \quad \|\overline{g}'_{i,l}-\overline{g}_{i,l}\|_{2} \in[\frac{2\sqrt{2}}{3}\delta C^{e}_{i,l}, \frac{4\sqrt{2}}{3}\delta C^{e}_{i,l}]$$
% $$\|h_{i, l}\|_{2} \in [\frac{2}{3}C^{f}_{i,l}, \frac{4}{3}C^{f}_{i,l}],\quad \|h'_{i, l}-h_{i, l}\|_{2} \in [\frac{2}{3}\delta  C^{e}_{i,l}, \frac{4}{3}\delta C^{e}_{i,l}]$$
$\|h_{i, l}\|_{2} \in [\frac{2}{3}C^{f}_{i,l}, \frac{4}{3}C^{f}_{i,l}]$.
\end{lemma2}

\newtheorem{lemma3}[theorem]{Lemma}
\begin{lemma3}
\label{A3}
If $m=\Omega(LlogL)$, for any fixed input $\boldsymbol{X}=\{x_{i} \in \boldsymbol{S}: i \in [n]\}$ , with probability $1-e^{-\Omega(m/L)}$ over the randomness of initialization, we have for every $l \in [L]$ and every $i \in [n]$,$\|a^{\top}\boldsymbol{D}_{i,L}\boldsymbol{W}_{i,L}\cdot\cdot\cdot \boldsymbol{D}_{i,l}\boldsymbol{W}_{i,l}\|_{2} = O(\frac{\sqrt{mL}}{\overline{C}^f_L})$.

% \color[rgb]{1,0,0}{the probability may need to be changed for choosing any i (?)}
\end{lemma3}
% \textit{Proof.}
% This is a restatement of Lemma A.1 in \cite{gao2019convergence}, since our GNN architecture shares the same weight initialization method as DNN.
\textit{Proof of lemma\ref{A1}, lemma\ref{A2}, lemma\ref{A3}.}
 These are restatements of Lemma A.1, Lemma A.2 and Lemma A.3 in \cite{gao2019convergence}. Since our GNN architecture shares the same weight initialization method as regular DNN, we can directly apply these results over the randomness of initialization.
In Lemma~\ref{A2} the value is different from Lemma A.2 in \cite{gao2019convergence} due to aggregation, but with the same random weight initialization, the probability and relative bound is consistent.

We prove the perturbation of gradients caused by perturbation of weights is small following \cite{allen2019convergence}.
%can be applied with minor change considering the scale of intermediate layer embedding caused by aggregation.

% lemma7
\newtheorem{lemma4}[theorem]{Lemma}
\begin{lemma4}
\label{A5}
Given any fixed inputs $\boldsymbol{X}=\{x_{i} \in \boldsymbol{S}: i \in [n]\}$. If $ m \ge max(d, \Omega(LlogL)) $, $\frac{R}{\sqrt{m}} \le \frac{1}{SL^6(\log m)^3}$ for some sufficiently large constant S, with probability $1-O(L)e^{-\Omega((mR)^{2/3}L)}$ over the randomness of initialization, we have for any $\boldsymbol{W} \in B(R)$ , any $l \in [L]$ and every $i \in [n]$,
$$
% \nabla
\|\frac{\partial f_{i}(\boldsymbol{W},X)}{\partial \boldsymbol{W}_l} - \frac{\partial f_{i}(\boldsymbol{W}^{(0)},X)}{\partial \boldsymbol{W}_l} \|_{F} = O(\frac{R^{1/3}m^{1/3}L^{2}\sqrt{\log m}}{\overline{C}^f_L}),
$$
$$
\|\frac{\partial f_{i}(\boldsymbol{W},X)}{\partial \boldsymbol{W}_l}\|_{F} = O(\frac{\sqrt{mL}}{\overline{C}^f_L})
$$

\end{lemma4}
% here $\hat{C}_{i,l} = \underbrace{\sum_{i_{L-1}\in N(i)}\frac{1}{|N(i)|}\cdot\cdot\cdot\sum_{i_l\in N(i_{l+1})}\frac{1}{|N(i_{l+1})|}} _{L-l\ sums}(C^f_{i,l-1})^{2/3}(C^{e}_{i, l})^{1/3}$. 
% This is the fundamental factor that determines how much rounding error a graph can tolerate.

\textit{Proof.}
We denote $\boldsymbol{D}'$, $\boldsymbol{W}'$, and $h'$ as perturbed $\boldsymbol{D}$, $\boldsymbol{W}$, and $h$, respectively. With Claim 8.3 and Lemma 8.2(c) of \cite{allen2019convergence}, we have that when $\frac{R}{\sqrt{m}} \le \frac{1}{SL^6(\log m)^3}$, with probability $1-e^{-\Omega(m(R/\sqrt{m})^{2/3}L)}$,
\begin{equation}
    \label{delta_D}
        \|\boldsymbol{D}'_{i,l}-\boldsymbol{D}_{i,l}\|_0 = O(m(\frac{R}{\sqrt{m}})^{2/3}L)
\end{equation}
and
\begin{equation}
    \label{delta_h}
        \|h'_{i,l}-h_{i,l}\|_0 = O(C^f_{i,l}\frac{R}{\sqrt{m}} L^{5/2}\sqrt{\log m})        
\end{equation}

Combining (\ref{delta_D}) and Lemma 8.7 of \cite{allen2019convergence}, we get:

\begin{align*}
& \|(a^{\top}\boldsymbol{D}'_{i,L}\boldsymbol{W}'_L \cdot\cdot\cdot \boldsymbol{D}'_{i_{l+1},l+1}\boldsymbol{W}'_{l+1}\boldsymbol{D}'_{i_L,l})-(a^{\top}\boldsymbol{D}_{i,L}\boldsymbol{W}_L \cdot\cdot\cdot \boldsymbol{D}_{i_{l+1},l+1}\boldsymbol{W}_{l+1}\boldsymbol{D}_{i_L,l})\|_{2}\\
& \le O(\frac{1}{\overline{C}^f_L}(\frac{R}{\sqrt{m}})^{1/3}L^{2}\sqrt{m\log m})\\
\end{align*}

During backward propagation, we have:
$$
\frac{\partial f_{i}(\boldsymbol{W},X)}{\partial \boldsymbol{W}_L}  = \underbrace{\sum_{i_{L-1}\in N(i)}\frac{1}{|N(i)|}\cdot\cdot\cdot\sum_{i_L\in N(i_{l+1})}\frac{1}{|N(i_{l+1})|}} _{L-l\ sums}
h_{i_L, l-1}a^{\top}\boldsymbol{D}_{i,L}\boldsymbol{W}_L \cdot\cdot\cdot \boldsymbol{D}_{i_{l+1},l+1}\boldsymbol{W}_{l+1}\boldsymbol{D}_{i_L,l}
$$
Then, using Lemma\ref{A2}, Lemma\ref{A3}, and the above results, we have:

\begin{align*}
&\|\frac{\partial f_{i}(\boldsymbol{W},X)}{\partial \boldsymbol{W}_l}  - \frac{\partial f_{i}(\boldsymbol{W}^{(0)},X)}{\partial \boldsymbol{W}_l} \|_{F}=
\| \underbrace{\sum_{i_{L-1}\in N(i)}\frac{1}{|N(i)|}\cdot\cdot\cdot\sum_{i_L\in N(i_{l+1})}\frac{1}{|N(i_{l+1})|}} _{L-l\ sums} \\
&h'_{i_L, l-1}a^{\top}\boldsymbol{D}'_{i,L}\boldsymbol{W}'_L\cdot\cdot\cdot \boldsymbol{D}'_{i_{l+1},l+1}\boldsymbol{W}'_{l+1}\boldsymbol{D}'_{i_L,l}-h_{i_L, l-1}a^{\top}\boldsymbol{D}_{i,L}\boldsymbol{W}_L \cdot\cdot\cdot \boldsymbol{D}_{i_{l+1},l+1}\boldsymbol{W}_{l+1}\boldsymbol{D}_{i_L,l} \|_{F}  \\
% ===
\le&\underbrace{\sum_{i_{L-1}\in N(i)}\frac{1}{|N(i)|}\cdot\cdot\cdot\sum_{i_L\in N(i_{l+1})}\frac{1}{|N(i_{l+1})|}} _{L-l\ sums}
 [\|h'_{i_L, l-1}-h_{i_L, l-1}\|_{2}\cdot\|a^{\top}\boldsymbol{D}_{i,L}\boldsymbol{W}_L \cdot\cdot\cdot \boldsymbol{W}_{l+1}\boldsymbol{D}_{i_L,l}\|_{2}\\ 
 &+\|h_{i_L, l-1}\|_{2}\cdot\|(a^{\top}\boldsymbol{D}'_{i,L}\boldsymbol{W}'_L \cdot\cdot\cdot \boldsymbol{W}'_{l+1}\boldsymbol{D}'_{i_L,l})-(a^{\top}\boldsymbol{D}_{i,L}\boldsymbol{W}_L \cdot\cdot\cdot \boldsymbol{W}_{l+1}\boldsymbol{D}_{i_L,l})\|_{2}\\
%  ===
\le&\underbrace{\sum_{i_{L-1}\in N(i)}\frac{1}{|N(i)|}\cdot\cdot\cdot\sum_{i_L\in N(i_{l+1})}\frac{1}{|N(i_{l+1})|}} _{L-l\ sums}
 [O(C^f_{i,l-1}\frac{R}{\sqrt{m}} L^{5/2}\sqrt{\log m})\cdot O(\frac{\sqrt{mL}}{\overline{C}^f_L})\\
 & +O(C^f_{i,l-1})\cdot O(\frac{1}{\overline{C}^f_L}(\frac{R}{\sqrt{m}})^{1/3}L^{2}\sqrt{m\log m})]\\
%  (?)&+(L-l)O(C^{f}_{i,l-1}\sqrt{mL}R)] \\
% ===
= &\underbrace{\sum_{i_{L-1}\in N(i)}\frac{1}{|N(i)|}\cdot\cdot\cdot\sum_{i_L\in N(i_{l+1})}\frac{1}{|N(i_{l+1})|}} _{L-l\ sums}
    % (L-l+2) \cdot C^{e}_{i_L, l}\cdot O(\delta\sqrt{mL})\\
    O(\frac{C^f_{i,l-1}}{\overline{C}^f_L} R^{1/3}m^{1/3}L^{2}\sqrt{\log m})\\
% ===
% = &\underbrace{\sum_{i_{L-1}\in N(i)}\frac{1}{|N(i)|}\cdot\cdot\cdot\sum_{i_L\in N(i_{l+1})}\frac{1}{|N(i_{l+1})|}} _{L-l\ sums}
%     % (L-l+2) \cdot C^{e}_{i_L, l}\cdot O(\delta\sqrt{mL})\\
%     % O(\delta C^{e}_{i,l-1}\sqrt{mL})\\
%     O( L^{3/2}(C^f_{i,l-1})^{2/3}(\delta C^e_{i,l})^{1/3}\sqrt{m\ log\ m})\\
% ===
% =& O(\hat{C}^{e}_{i,l}\delta\sqrt{mL})
%  =&O(\hat{C}^f_{i,l}\omega^{1/3}L^{2}\sqrt{m\log m})\\
 =&O(\frac{R^{1/3}m^{1/3}L^{2}\sqrt{\log m}}{\overline{C}^f_L})
\end{align*}
where $C^f_{i,l-1} \le 1$.

With Lemma\ref{A2}, Lemma\ref{A3}, (\ref{delta_h}), and $\frac{R}{\sqrt{m}} \le \frac{1}{SL^6(\log m)^3}$, we have
$$
\|\frac{\partial f_{i}(\boldsymbol{W},X)}{\partial \boldsymbol{W}_l}\|_{F} = O(\frac{\sqrt{mL}}{\overline{C}^f_L})\cdot (O(C^{f}_{i,l}) + O(C^f_{i,l}\frac{R}{\sqrt{m}} L^{5/2}\sqrt{\log m}) ) = O(\frac{C^{f}_{i,l}\sqrt{mL}}{\overline{C}^f_L}) = O(\frac{\sqrt{mL}}{\overline{C}^f_L})
$$

% here we select
% $$
% \frac{m}{logm} \ge \frac{(C^f_{i,l-1})^2L^2}{\delta C^e_{i,l-1}}
% $$

% \color[rgb]{1,0,0}{TBD.}
Then we are able to prove the loss function is almost convex within the neighborhood $B(R)$ for any fixed $X$.

% Lemma 8
\newtheorem{lemma5}[theorem]{Lemma}
\begin{lemma5}
\label{A6}
If $R = O(\frac{\sqrt{m}}{L^6(\log m)^3})$, with probability at least $1-O(nL)e^{-\Omega((mR)^{2/3}L)}$ over random initialization, we have for any $ \boldsymbol{W}^{(1)},\boldsymbol{W}^{(2)} \in B(R)$, any $\boldsymbol{X}=\{x_{i} \in \boldsymbol{S}: i \in [n]\}$, and any $Y = (y_1,\cdot\cdot\cdot,y_n) \in \mathbb{R}^n$,
\begin{align*}
l(f_{i}(\boldsymbol{W}^{(2)},X),y_{i}) &\ge l(f_{i}(\boldsymbol{W}^{(1)},X),y_{i}) + \left< \nabla_{\boldsymbol{W}} l(f_{i}(\boldsymbol{W},X), y_{i}), \boldsymbol{W}^{(2)}-\boldsymbol{W}^{(1)}\right>\\
&- \|\boldsymbol{W}^{(2)}-\boldsymbol{W}^{(1)}\|_{F}O(\frac{R^{1/3}m^{1/3}L^{5/2}\sqrt{\log m}}{\overline{C}^f_{L}})
\end{align*}
\end{lemma5}

\textit{Proof.}
For any fixed $X$ and any fixed $i$, with probability $1-O(nL)e^{-\Omega((mR)^{2/3}L)}$, we have:
\begin{align*}
&l(f_{i}(\boldsymbol{W}^{(2)},X),y_{i}) - l(f_{i}(\boldsymbol{W}^{(1)},X),y_{i}) - \left< \nabla_{\boldsymbol{W}} l(f_{i}(\boldsymbol{W},X), y_{i}), \boldsymbol{W}^{(2)}-\boldsymbol{W}^{(1)}\right>\\
& \ge \frac{\partial}{\partial f}l(f_{i}(\boldsymbol{W}^{(1)},X), y_{i})[f_{i}(\boldsymbol{W}^{(2)}, X)-f_{i}(\boldsymbol{W}^{(1)}, X) - \left< \nabla _{\boldsymbol{W}}f_{i}(\boldsymbol{W}^{(1)},X), \boldsymbol{W}^{(2)}-\boldsymbol{W}^{(1)} \right>]\\
& = \frac{\partial}{\partial f}l(f_{i}(\boldsymbol{W}^{(1)},X), y_{i})\left< \int_{0}^{1}(\nabla_{\boldsymbol{W}}f_{i}(t\boldsymbol{W}^{(2)}+(1-t)\boldsymbol{W}^{(1)}, X) -   \nabla _{\boldsymbol{W}}f_{i}(\boldsymbol{W}^{(1)},X))dt, \boldsymbol{W}^{(2)}-\boldsymbol{W}^{(1)} \right>\\
&\ge -\|\boldsymbol{W}^{(2)}-\boldsymbol{W}^{(1)}\|_{F}O(\frac{R^{1/3}m^{1/3}L^{5/2}\sqrt{\log m}}{\overline{C}^f_{L}})\\
% & = -\|\boldsymbol{W}^{(2)}-\boldsymbol{W}^{(1)}\|_{F}O(\hat{C}_{i}(mR)^{1/3}L^{2}\sqrt{log \ m})\\
\end{align*}
The first inequality is due to convexity of $l$ with regard to $f$ and the second inequality is due to Lemma \ref{A5}. $\frac{\partial l}{\partial f}$ is bounded and $\frac{\partial f}{\partial \boldsymbol{W}} = (\frac{\partial f}{\partial \boldsymbol{W}_1}, \cdot\cdot\cdot, \frac{\partial f}{\partial \boldsymbol{W}_L})$.

\subsection{Loss after Quantization}
\label{similar_loss}               
In this subsection we prove with the optimal weights of uncompressed features, the difference between losses using quantized features and original features as input is small.

\newtheorem{lemma6}[theorem]{Lemma}
\begin{lemma6}
\label{close_loss}
Let $\boldsymbol{W}^{(*)}=\arg\min_{\boldsymbol{W}\in B(R)}Loss(\boldsymbol{W},X)$ be the optimal weights of GNN trained using original features $X$, with loss function satisfying Assumption\ref{smooth_loss}, use the same weights on quantized features $X'$ satisfying $\|x_i'-x_i\|_2 \le \delta$ and Assumption \ref{iid_error}, with high probability over the randomness of initialization of $a^\top$ we have
$$
Loss(\boldsymbol{W}^{(*)}, X')- Loss(\boldsymbol{W}^{(*)}, X) = O(\hat{C}_L\delta)
$$
\end{lemma6}
\textit{Proof.}
We denote $\boldsymbol{W}^{(*)}$ as the optimal weights without feature compression, so the optimal loss is $Loss(\boldsymbol{W}^{(*)}, X)$. We use $h_{i,l}(X)$ to indicate $h_{i,l}$ with regard to input feature $X$, we can get 
\begin{align*}
Loss(\boldsymbol{W}^{(*)}, X')- Loss(\boldsymbol{W}^{(*)}, X) =& \frac{1}{n}\sum_{i=1}^{n}[l(f_i(\boldsymbol{W}^{(*)}, X'), y_i)-l(f_i(\boldsymbol{W}^{(*)}, X), y_i)]\\
    = & \frac{1}{n}\sum_{i=1}^{n}\frac{\partial l}{\partial f}(f_i(\boldsymbol{W}^{(*)}, X')-f_i(\boldsymbol{W}^{(*)}, X))\\
    = & \frac{1}{n}\sum_{i=1}^{n}\frac{\partial l}{\partial f}(a^\top (h_{i,L}(X')-h_{i,L}(X)))\\
    % \le & \frac{1}{n}\sum_{i=1}^{n}\frac{\partial l}{\partial f}\|a\|_2 \cdot \|h'_{i,L}-h_{i,L}\|_2)\\
    = & \frac{1}{n}\sum_{i=1}^{n}O(\frac{1}{\overline{C}^f_{L}})\cdot O(\delta C^e_{i,L})\\
    = & O(\frac{\delta \overline{C}^e_{L}}{\overline{C}^f_{L}})\\
    = & O(\hat{C}_L\delta)
    % = & O(R \overline{C}^{e}_{L})\\
\end{align*}
where we define $\hat{C}_L = \frac{\overline{C}^e_{L}}{\overline{C}^f_{L}}$.

\subsection{Proof of Theorem\ref{convergence}}
\label{final_proof}

With Lemma\ref{A5}, Lemma\ref{A6}, and Lemma\ref{close_loss}, we are ready to prove Theorem\ref{convergence}.

\textit{Proof of Theorem\ref{convergence}.}
For a projected gradient descent with in total $T$ steps starting from initialization $\boldsymbol{W}^{(0)}$, we denote $\boldsymbol{W}^{(t)}$ as the weights after $t$ steps with step size $\alpha$.
$\boldsymbol{W}^{(t)} \in B(R)$ holds for all $ t = 0, 1, \cdot\cdot\cdot, T$. 

The update rule of projected gradient descent is $\boldsymbol{W}^{(t+1)} = \mathcal{P}_{B(R)}(\boldsymbol{V}^{(t+1)})$, $\boldsymbol{V}^{(t+1)} = \boldsymbol{W}^{(t)} - \alpha \nabla_{\boldsymbol{W}}Loss(\boldsymbol{W}^{(t)})$.
Let $d_{t} = \|\boldsymbol{W}^{(t)} - \boldsymbol{W}^{(*)}\|_{F}$. If $R = O(\frac{\sqrt{m}}{L^6(\log m)^3})$ and loss function satisfies Assumption\ref{smooth_loss}, with probability at least $1-O(nL)e^{-\Omega((mR)^{2/3}L)}$ over random initialization, we have 
\begin{align*}
d_{t+1}^{2} = & \|\boldsymbol{W}^{(t+1)} - \boldsymbol{W}^{(*)}\|_{F}^{2}\\
    \le & \|\boldsymbol{V}^{(t+1)} - \boldsymbol{W}^{(*)}\|_{F}^{2}\\
    = & \|\boldsymbol{W}^{(t)}-\boldsymbol{W}^{(*)}\|_{F}^{2} + 2\left< V_{t+1}-\boldsymbol{W}^{(t)}, \boldsymbol{W}^{(t)}-\boldsymbol{W}^{(*)} \right> + \|V_{t+1} - \boldsymbol{W}^{(t)}\|_{F}^{2}\\
    = & d_{t}^{2} + 2\alpha \left<\nabla_{\boldsymbol{W}}L'(\boldsymbol{W}^{(t)}), \boldsymbol{W}^{(*)}-\boldsymbol{W}^{(t)} \right> + \alpha^{2}\|\nabla_{\boldsymbol{W}}L'(\boldsymbol{W}^{(t)})\|_{F}^{2}\\
    = & d_{t}^{2} + \frac{2\alpha}{n} \sum_{i=1}^{n}\left<\nabla_{\boldsymbol{W}}l(f_{i}(\boldsymbol{W}^{(t)},X'),y_{i}), \boldsymbol{W}^{(*)}-\boldsymbol{W}^{(t)} \right> + \alpha^{2}\|\frac{1}{n}\sum_{i=1}^{n}\frac{\partial l}{\partial f}\nabla_{\boldsymbol{W}}f_{i}(\boldsymbol{W}^{(t)},X')\|_{F}^{2}\\
    \le & d_{t}^{2} + \frac{2\alpha}{n} \sum_{i=1}^{n}
    [l(f_{i}(\boldsymbol{W}^{(*)},X'),y_{i}) - l(f_{i}(\boldsymbol{W}^{(t)},X'),y_{i}) + \\ 
    & \|\boldsymbol{W}^{(*)}-\boldsymbol{W}^{(t)}\|_{F}O(\frac{R^{1/3}m^{1/3}L^{5/2}\sqrt{\log m}}{\overline{C}^f_{L}})] + \alpha^{2}O(\frac{mL^{2}}{(\overline{C}^{f}_L)^{2}})\\
    % = & d_{t}^{2} + \frac{2\alpha}{n} \sum_{i=1}^{n}[l(f_{i}(\boldsymbol{W}^{(*)},X'),y_{i}) - l(f_{i}(\boldsymbol{W}^{(t)},X'),y_{i})] + 
    % O(\alpha\frac{R^{4/3}m^{-1/6}L^{5/2}\sqrt{\log m}}{\overline{C}^f_{L}}+\alpha^{2}mL^{2})\\
    \le & d_{t}^{2} + 2\alpha [Loss(\boldsymbol{W}^{(*)},X) + O(\hat{C}_L\delta) - Loss(\boldsymbol{W}^{(t)},X')] + \\
    & O(\alpha\frac{R^{4/3}m^{-1/6}L^{5/2}\sqrt{\log m}}{\overline{C}^f_{L}}+\alpha^{2}\frac{mL^{2}}{(\overline{C}^{f}_L)^{2}})\\
    \le & d_{t}^{2} + 2\alpha [Loss(\boldsymbol{W}^{(*)},X) - Loss(\boldsymbol{W}^{(t)},X')] + 
    O(\alpha\frac{R^{4/3}m^{-1/6}L^{5/2}\sqrt{\log m}}{\overline{C}^f_{L}}+\alpha^{2}\frac{mL^{2}}{(\overline{C}^{f}_L)^{2}} + \alpha\hat{C}_L\delta)\\    
    % \le & d_{t}^{2} + 2\alpha [Loss(\boldsymbol{W}^{(*)},X) - Loss(\boldsymbol{W}^{(t)},X')] + 
    % O(\alpha\hat{C}_Lm^{-1/6}R^{4/3}L^{2}\sqrt{log \ m}+\alpha^{2}(\overline{C}^{f}_L)^{2}mL^{2}+\alpha  \delta \overline{C}^{e}_{L})\\    
\end{align*}
where the second inequality is for Lemma\ref{A6} and Lemma\ref{A5}, the third inequality is due to Lemma\ref{close_loss}. Note that Lemma\ref{A6} and Lemma\ref{A5} can be satisfied with $m=max(\Omega(\frac{L^{16}R^{9}}{(\overline{C}^f_L)^7\epsilon^7}), \Omega(d^2))$.

Using induction, we have
\begin{align*}
d_{T}^{2} \le & d_{0}^{2} + 2\alpha\sum_{t=0}^{T+1}[Loss(\boldsymbol{W}^{(*)},X) - Loss(\boldsymbol{W}^{(t)},X')] + \\
& O(T(\alpha\frac{R^{4/3}m^{-1/6}L^{5/2}\sqrt{\log m}}{\overline{C}^f_{L}}+\alpha^{2}\frac{mL^{2}}{(\overline{C}^{f}_L)^{2}} + \alpha\hat{C}_L\delta))
\end{align*}
which implies that 
\begin{align*}
\min_{0\le t \le T}(Loss(\boldsymbol{W}^{(t)},X') - Loss(\boldsymbol{W}^{(*)},X)) \le & \frac{d_{0}^{2}-d_{T}^{2}}{\alpha T}+O(\frac{R^{4/3}m^{-1/6}L^{5/2}\sqrt{\log m}}{\overline{C}^f_{L}}+\alpha \frac{mL^{2}}{(\overline{C}^{f}_L)^{2}} + \hat{C}_L\delta)\\
\le & \frac{R^{2}}{m\alpha T} +O(\frac{R^{4/3}m^{-1/6}L^{5/2}\sqrt{\log m}}{\overline{C}^f_{L}}+\alpha \frac{mL^{2}}{(\overline{C}^{f}_L)^{2}} + \hat{C}_L\delta)\\
% \le & \frac{R^{2}}{m\alpha T} +O(\delta\hat{C}^{e}_{i,l}LR+\alpha(\overline{C}^{f}_L)^{2}mL^{2}+R \overline{C}^e_{i,L})\\
\le & \epsilon
\end{align*}

Where the last inequality is due to parameter selection $\alpha=O(\frac{\epsilon(\overline{C}^{f}_L)^{2}}{mL^2})$, $T=\Theta(\frac{R}{m\alpha \epsilon})$, $m=\Omega(\frac{L^{16}R^{9}}{(\overline{C}^f_L)^7\epsilon^7})$ and $\delta=O(\frac{\epsilon}{\hat{C}_L})$. 

Given $\epsilon>0$, suppose $R=\Omega(1)$ and $m=max(\Theta(\frac{L^{16}R^{9}}{(\overline{C}^f_L)^7\epsilon^7}), \Theta(d^2))$. Let the loss function satisfy Assumption\ref{smooth_loss}. If we run projected gradient descent based on the convex constraint set $B(R)$ with step size $\alpha=O(\frac{\epsilon(\overline{C}^{f}_L)^{2}}{mL^2})$ for $T=\Theta(\frac{R}{m\alpha \epsilon})$ steps,  with high probability we have 
$$
\min_{0\le t \le T}(Loss(\boldsymbol{W}^{(t)},X') - Loss(\boldsymbol{W}^{(*)},X)) \le \epsilon
$$
where $\boldsymbol{W}^{(*)}=\arg\min_{\boldsymbol{W}\in B(R)}Loss(\boldsymbol{W},X)$.

The selection of $m$, $\alpha$, and $\delta$ within the above analysis relies on factors decided by graph properties. %%CL \cheng{Therefore, we xxx.} 
For simplicity we can directly use the bounds of these factors so that $m=max(\Omega(\frac{L^{16}R^{9}n^7}{\epsilon^7}), \Omega(d^2))$, $\alpha=O(\frac{\epsilon}{mL^2n^2})$, and $\delta=O(\epsilon)$. The detailed analysis of these factors are in Section \ref{factor_analysis}.

% Analysis TBD.
\subsection{Impacts of Graph Structure and Model Depth on Compression Ratios}
\label{factor_analysis}

To train a GNN model over a large graph data set, the key to using feature quantization is to choose a proper compression ratio to strike a balance between the saved memory and PCIe bandwidth consumption and the accuracy. 
With the previous analysis of loss bound, combined with our experiments, we identify that the major contributor to the loss bound (see Theorem~\ref{convergence}) is $\delta\hat{C}_L$. In particular, given an expected loss bound $\epsilon$,  the factor $\hat{C}_L$ is inversely proportional to $\delta$, where $\delta$ is roughly $\Theta(2^{-32/CR})$.  
% So we would expect the compression ratio to be $CR = \frac{32}{\log_2\hat{C}_L-\log_2\epsilon}$.
Therefore, we would expect the compression ratio to be negatively correlated to the $\hat{C}_L$. \textit{Since $\hat{C}_L$ is relevant to both model depth, i.e., number of layers $L$, and the structure of the graph data set, we further study the impacts of the two factors on the selection of compression ratios.}

%To guide the selection of compression ratio, it is important to study these two factors.
% The following analysis demonstrates the advantage of applying feature quantization to GNN training.
\noindent\textbf{Model depth.} We begin our analysis with understanding the impact of number of GNN layers.
We test the values of $\overline{C}^f_{L}$ and $\overline{C}^e_{L}$ with the different numbers of layers on Reddit dataset.
As expected, the values of $\overline{C}^f_{L}$ , $\overline{C}^e_{L}$, and $\hat{C}_L$ decrease when we increase the number of layers, as shown in  Table\ref{depth_factor}. 
The decrease in $\hat{C}_L$ is mainly because  $\overline{C}^f_{L}$ decreases much slower than $\overline{C}^e_{L}$.

\begin{table}[!t]
\small
    \caption{Impacts of model depth (the number of layers). $\hat{C}_L$ is the factor deciding acceptable compression ratio (smaller is better). %%CL \yanc{make all tables self-contained by giving the meaning of each factor.}
    }
    \label{depth_factor}
    \centering
    \begin{tabular}{ lccccc } 
        \toprule
         & \multicolumn{5}{c}{Number of layers (L)}    \\
        Factor & 2 & 3 & 5 & 10 & 20  \\
        \midrule
        $\overline{C}^f_{L}$    & 0.1216    & 0.1095    & 0.0983    & 0.0804    & 0.0691      \\
        $\overline{C}^e_{L}$    & 0.0152    & 0.0086    & 0.0047    & 0.0026    & 0.0018      \\        
        $\hat{C}_L$             & 0.1248    & 0.0788    & 0.0482    & 0.0325    & 0.0258      \\        
        \bottomrule
    \end{tabular}
\end{table}
\noindent\textbf{Graph structure.} Then, to examine the impact of graph structure, we artificially create graph data sets with different structures based on the Reddit data set. Basically, we keep the Reddit nodes but delete some of the edges with three different selection methods to simulate graphs with varied sparsity, resulting in three types of data sets, namely, RANDOM, CENTRALIZED, and UNIFORM. In more detail, RANDOM deletes randomly selected edges, and thus keeps the power-law distribution of node degrees. CENTRALIZED deletes edges between low-degree nodes with high priority, leading the edges in the revised graph to be more centralized on high-degree nodes than that of the original graph. In contrast, UNIFORM prioritizes to delete edges between high-degree nodes, and make the resulting graph to follow an almost uniform degree distribution.

\begin{table}[t]
\small
    \caption{Impact of graph structures. CR means maximum acceptable compression ratio.}
    \label{structure_factor}
    \centering
    \begin{tabular}{ llcccccc } 
        \toprule
        Method to& &  \multicolumn{6}{c}{Percentage of remaining edges.}    \\
        extract graph & & 10\% & 30\% & 50\% & 70\% & 90\% & 100\%  \\
        \midrule
        \multirow{5}{*}{RANDOM}     &   $\overline{C}^f_{L}$    & 0.1329    & 0.1141    & 0.1147    & 0.1116    & 0.1050    & 0.1098    \\
                                    &   $\overline{C}^e_{L}$    & 0.0356    & 0.0171    & 0.0127    & 0.0105    & 0.0092    & 0.0088    \\        
                                    &   $\hat{C}_L$             & 0.2677    & 0.1500    & 0.1109    & 0.0940    & 0.0875    & 0.0799    \\    
                                    % &   Original Acc(\%)        & 85.92     & 93.54     & 95.29     & 95.55     & 95.90     & 96.03     & 96.15    \\   
                                    % &   BiFeat Acc(\%)          & 85.06     & 93.40     & 95.00     & 95.48     & 95.87     & 95.92     & 96.12    \\  
                                    &   CR                      & 16        & 16        & 32        & 32        & 32        & 32    \\
        \midrule
        \multirow{5}{*}{\specialcell{CENTRALIZED}}   &   $\overline{C}^f_{L}$   & 0.4013    & 0.3789    & 0.3427    & 0.2836    & 0.1854    &     \\
                                    &   $\overline{C}^e_{L}$    & 0.3873    & 0.3423    & 0.2924    & 0.2166    & 0.0913    &   \\        
                                    &   $\hat{C}_L$             & 0.9650    & 0.9034    & 0.8548    & 0.7638    & 0.4923    &   \\    
                                    % &   Original Acc(\%)        & 71.70     & 72.49     & 73.86     & 76.31     & 79.94     & 89.19     &     \\        
                                    % &   BiFeat Acc(\%)          & 67.29     & 67.81     & 70.21     & 73.51     & 78.33     & 87.91     &     \\   
                                    &   CR                      & 4         & 4         & 8         & 16        & 16        &     \\
        \midrule
        \multirow{5}{*}{\specialcell{UNIFORM}}    &   $\overline{C}^f_{L}$    & 0.1109    & 0.1111    & 0.1113    & 0.1022    & 0.1134    &    \\
                                    &   $\overline{C}^e_{L}$    & 0.0169    & 0.0104    & 0.0089    & 0.0086    & 0.0087    &   \\        
                                    &   $\hat{C}_L$             & 0.1523    & 0.0938    & 0.0802    & 0.0845    & 0.0770    &    \\    
                                    % &   Original Acc(\%)        & 93.02     & 96.44     & 96.24     & 96.25     & 96.17     & 96.23     &     \\        
                                    % &   BiFeat Acc(\%)          & 93.09     & 96.33     & 96.24     & 96.23     & 96.19     & 96.13     &     \\          
                                    &   CR                      & 32        & 32        & 32        & 32        & 32        &    \\
        \bottomrule
    \end{tabular}
\end{table}

%%CL We compare graphs with different average degrees and with the same average degrees but different distribution of edges.
Table \ref{structure_factor} shows the values of $\hat{C}_L$ with different graph structures under varied sparsity. Here, sparsity relates to the number of edges, i.e., more edges, denser graph, and vice versa. We also list the values of $\overline{C}^f_{L}$ and $\overline{C}^e_{L}$, as $\hat{C}_L = \overline{C}^e_{L} / \overline{C}^f_{L} $. Furthermore, $CR$ denotes the maximum compression ratio with less than 0.1\% accuracy degradation. Note that, here, we use BiFeat-SQ for the analysis, where the maximum CR is 32. 

We draw the following conclusions from Table~\ref{structure_factor}. First, across the three data sets, denser graphs (higher percentage of remaining edges, in comparison to the original Reddit data set) always have smaller $\hat{C}_L$. Therefore, their compression ratios can be higher. With the same sparsity (the same percentage of remaining edges), the graphs created by the CENTRALIZED method have the following properties: some nodes are isolated, and the remaining nodes are connected by several super nodes. The direct consequence is to have a higher $\hat{C}_L$ and low acceptable compression ratio. Contrary, UNIFORM has a lower $\hat{C}_L$ and higher compression ratio. The reasons are as follows. The existence of super nodes makes the quantization errors spread to a large amount of nodes, which however can hardly be cancelled out during the aggregation, because neighbors of each node mainly have this component. But, the UNIFORM method addresses this limitation by eliminating super nodes and keeping each node's neighborhoods less overlapped. %%CL , and thus helps cancel out the quantization error. 

%%CL Here, we exercise a few examples to walk through the above analysis. First, as an extreme case, we consider a graph with all nodes isolated. In such case, the GNN model looks similar to DNN, and thus has $\hat{C}_L=1$, as defined in \cite{allen2019convergence}. However, when considering Reddit graph, a three-layer GNN model has $\hat{C}_L\approx0.08$. This implies that with sufficient width, GNN can have a similar performance with 12$\times$ larger quantization error. With $CR$ being the compression ratio, the quantization error $\Theta(2^{-32/CR})$.  So 1-bit scalar quantization in GNN is similar in performance with 5-bit in DNN, while having 5$\times$ compression ratio.   

In summary, the above results indicate that BiFeat training is especially useful to handle deeper GNNs and graphs with degree distribution that is not very centralized. %%CL In the cases where higher compression ratios can be used, BiFeat-VQ shows the most promising results. 
%%CL In sparse or very centralized graphs, a lower compression ratio usually yields better results.

\section{Detailed Setups of Experiments}
In this section, we show the detailed setups corresponding to experiments presented in Section~\ref{sec:evaluation}.

\subsection{Accuracy Validation}
\noindent\textbf{Node Property Prediction.} 
For GraphSAGE, GAT, and ClusterGCN, we use the example code from DGL\cite{DGL}. We changed GAT implementation to make it support neighbor sampling and able to train on giant graphs.

ClusterGCN's default implementation uses the subgraph of train set nodes for training, which has poor performance in OGBN-Papers100M. 
We change this implementation by using the subgraph including train set nodes' 1-hop neighbors. 

For MLP model, we use a three-layer MLP with ReLU activation and dropout, where its hidden layer dimension is 256. 
The SQ and VQ setups are

The detailed hyperparameters are listed in Table \ref{node_acc_setup}. 
Not listed hyperparameters use the default value.
% in SQ or VQ, it means that it uses the same value as Full.

The SQ parameter means the BiFeat-SQ setup used in the test. Here all tests are done with this parameter as 1, meaning we only use the sign bit to indicate the original float32 number.
The VQ parameter includes two numbers, $width$ and $length$, are shown with the format $width$ - $length$. These parameters mean the number of dimensions of each codebook entry and the number of entries in each codebook, they will be discussed in detail in Appendix\ref{code_setup}.

\begin{table}[!t]
\small
    \caption{Experimental setups for the node property prediction tasks}
    \label{node_acc_setup}
    \centering
    \begin{tabular}{ llccc } 
        \toprule
        \multicolumn{2}{c}{} & \multicolumn{3}{c}{Datasets}  \\
        Model & Hyperparameters& Reddit   & OGBN-Papers100M & MAG240M  \\
        \midrule
        \multirow{5}{*}{GraphSAGE}  & \#layer       &  2            &   3           &   3\\
                                    & \#hidden      &  64           &   256         &   256\\
                                    & fan out       &  10,25        &   5,10,15     &   5,10,15\\
                                    % & lr            &  0.003        &   0.003       &   0.003\\                                    
                                    & SQ            &  1            &   1           &   1\\
                                    & VQ            &  100-16384    &   16-2048     &   16-2048\\        
        \midrule
        \multirow{6}{*}{GAT}        & \#layer       &  2            &   3           &   3\\
                                    & \#hidden      &  64           &   256         &   256\\
                                    & fan out       &  10,25        &   5,10,15     &   5,10,15\\
                                    & drop out      &  0.2          &   0.25        &   0.25\\                                    
                                    % & batch size    &  1000         &   1000        &   1000       \\
                                    & lr            &  0.003        &   0.005       &   0.001\\                                      
                                    & SQ            &  1            &   1           &   1\\
                                    & VQ            &  51-1024      &   16-16384    &   16-2048\\ 
        \midrule
        \multirow{5}{*}{ClusterGCN} & \#hidden      &  128          &   256         &   -\\
                                    & psize         &  1000         &   1000        &   -\\
                                    & batch size    &  100          &   20          &   -\\
                                    & dropout       &  0.25         &   0.1         &   -\\  
                                    & lr            &  0.05         &   0.001       &   -\\                                     
                                    & SQ            &  1            &   1           &   -\\
                                    & VQ            &  37-256       &   16-2048     &   -\\         
        \midrule
        \multirow{4}{*}{MLP}        & \#layer       &  3            &   3           &   3\\
                                    & \#hidden      &  256          &   256         &   256\\
                                    % & batch size    &  1000         &   1000        &   1000\\
                                    % & dropout       &  0.5          &   0.5         &   0.5\\  
                                    % & lr            &  0.001        &   0.001       &   0.001\\                                     
                                    & SQ            &  1            &   1           &   1\\
                                    & VQ            &  16-1024      &   16-8192     &   12-2048\\  
        \bottomrule
    \end{tabular}
\end{table}

\noindent\textbf{Link Property Prediction.}
We test GCN and GraphSAGE with the OGB example code. Two models both use the default setting.
The detailed setups are shown in Table\ref{link_acc_setup} and Table\ref{graph_acc_setup}, with the same manner as node property prediction setups.

\begin{table}[!t]
\small
    \caption{Experimental setups for the link property prediction tasks}
    \label{link_acc_setup}
    \centering
    \begin{tabular}{ llcc } 
        \toprule
        \multicolumn{2}{c}{} & \multicolumn{2}{c}{Datasets}  \\
        Model & Hyperparameters  & OGBL-COLLAB & OGBL-PPA  \\
        \midrule
        \multirow{5}{*}{GCN}        & \#layer       &  3            &   3           \\
                                    & \#hidden      &  256           &   256        \\
                                    & lr            &  0.001        &   0.01    \\                                        
                                    & SQ            &  1            &   1           \\
                                    & VQ            &  16-2048    &   16-2048     \\        
        \midrule
        \multirow{6}{*}{GraphSAGE}  & \#layer       &  3            &   3           \\
                                    & \#hidden      &  256           &   256        \\
                                    & lr            &  0.001        &   0.01    \\                                        
                                    & SQ            &  1            &   1           \\
                                    & VQ            &  16-2048    &   16-2048     \\  
        \bottomrule
    \end{tabular}
\end{table}
\noindent\textbf{Graph Property Prediction.}
We used the DGL example code of GIN and added GCN implementation. In our tests, GCN uses the same setup as GIN.
The detailed setups are shown in Table\ref{graph_acc_setup}. Due to the original features being one-hot codes, we didn't test BiFeat-SQ, and we used a much more aggressive BiFeat-VQ setup. 

\begin{table}[!t]
\small
    \caption{Experimental setups for the graph property prediction tasks}
    \label{graph_acc_setup}
    \centering
    \begin{tabular}{ llcccc } 
        \toprule
        \multicolumn{2}{c}{} & \multicolumn{4}{c}{Datasets}  \\
        Model & Hyperparameters  & PTC & MUTAG & PROTEINS & COLLAB \\
        \midrule
        \multirow{5}{*}{GCN}        & \#GNN layer       &  5            &   5   &    5   &   5        \\
                                    & \#MLP layer       &  2            &   2   &    2   &   2        \\
                                    & \#hidden      &  64           &   64          &  64           &   64      \\
                                    & degree as label   & false     & false      & false    & true\\
                                    & graph pooling type & sum  & sum   & sum   & mean  \\
                                    & VQ            &  19-16    &   7-4     &   17-16   &  367-256      \\        
        \midrule
        \multirow{5}{*}{GIN}        & \#GNN layer       &  5            &   5   &    5   &   5        \\
                                    & \#MLP layer       &  2            &   2   &    2   &   2        \\
                                    & \#hidden      &  64           &   64          &  64           &   64      \\
                                    & degree as label   & false     & false      & false    & true\\
                                    & graph pooling type & sum  & sum   & sum   & mean  \\
                                    & VQ            &  19-16    &   7-4     &   17-16   &  367-256      \\        
        \bottomrule
    \end{tabular}
\end{table}

\subsection{Accleration and Giant Graph}
The tests of training acceleration and giant graph support use the same hyperparameter setup as the accuracy validation. 
In these tests, unless stated otherwise,  we use a single card for training, and 5-7 CPU workers do sampling, making sure sampling is not the bottleneck.

We use the code of PaGraph\cite{lin2020pagraph} to test the performance of caching. We made minor modifications to make it work efficiently in the newer DGL version and kept its static caching strategy. We reserve about 1GB of GPU memory and use all other free memory for cache. As a result, for OGBN-Papers100M, about 8\% original features or all quantized features can be cached; for MAG240M, about 0.5\% of original features or 12\% of the quantized features can be cached.

For the MAG240M dataset, we use the example code from OGB to do preprocessing, generate authors' features by averaging their papers' features, and generate institutions' features with authors' features. At last, we convert the heterogeneous graph into an undirected homogeneous graph.

During the training of GraphSAGE and GAT on the MAG240M dataset with original features, the peak memory usage is about 420GB. Using multiple cards for training in this situation almost used up CPU resources and significantly increased sampling time cost. Increasing the number of CPU sampling workers won't help. This problem didn't happen when using BiFeat, due to peak memory usage being about 70GB and only about half of the CPU resources being used.

\section{Open Source Code and Installation Instructions}
\label{code_setup}
We publish our BiFeat code anonymously on GitHub at  \underline{https://github.com/BiFeat/BiFeat}. In addition, we provide a guideline for the hyperparameter tuning of BiFeat.

\noindent\textbf{BiFeat-SQ.} 
There is only one parameter needed to configure, that is, the number of bits used to indicate a number. Setting this to 1 bit is practical for most cases, where only positive or not matters.
%meaning we don't care about the absolute value of the number, only it's positive or not is important. 
For other cases, 2-bit is usually enough. Though there are other setups like the percentage of long-tailed values to be clipped, they are not very important in such a low bit width. 

BiFeat-SQ is also efficient as it needs no more than several minutes to quantize even the largest MAG240M dataset.

\noindent\textbf{BiFeat-VQ.} 
Recall that the theoretical compression ratio of BiFeat-VQ is $\frac{width*32}{\log_2length}$, where $length$ is the number of entries in each codebook and $width$ is the dimension of each entry. There are $\lceil \frac{\#feature\ dimension}{width}\rceil $ codebooks in total. 
Here we do not count the size of codebooks for two reasons. First, its size is small compared with features. Second, it is always kept in GPU. Thus we do not need to move it between CPU and GPU memory. 

We can see that $length$ has less impact on the compression ratio. In our experiments, increasing $length$ yields less accuracy degradation with almost the same compression ratio.  
However, a large $length$ would significantly increase the time cost of vector quantization. For MAG240M, it may take up to a whole day to perform compression. This is because BiFeat-VQ needs to calculate the pair-wise distance to choose a proper entry in the codebook for each node, and thus the time cost is proportional to the $length$. We observe the acceptable $length$ is at most 16384. Thus each vector needs 2 bytes to store the index of the codebook entry due to bit alignment. Without a highly complex method to pack bits, we advise using a $length$ of 256 for convenience, paired with a width capable of reaching the optimal accuracy while as big as possible. This would also save the pre-processing time, sacrificing maximum compression ratio.
\section{Discussions on Time Cost of Feature Quantization}
We did not fully discuss the pre-processing time for quantizing feature data in our submission. This is because it depends on the compression method and the choice of compression parameters. For BiFeat-SQ, it is very efficient. Even when processing the MAG240M dataset (up to 375GB), it takes no more than 15 minutes using the highest compression rate. However, for BiFeat-VQ, it takes hours to process the MAG240M dataset because of the calculation of the global information. To accelerate BiFeat-VQ pre-processing, we uniformly sample a portion of the data instead of global data for statistics. The uniform random sampling strategy we use may not be optimal as the features are accessed unevenly in GNN training. We plan to explore other strategies in our future work. In addition, it is more efficient to use the GPU to accelerate BiFeat-VQ pre-processing. We also leave this as our future work.
%have not made any attempts in this regard.

\end{document}